\documentclass[lettersize,journal]{IEEEtran}

\usepackage{xspace}             % 必须先加载，否则 \xspace 无效
\usepackage{amsmath,amsfonts}   % \eqref 来自 amsmath
\usepackage{algorithmic}
\usepackage{algorithm}
\usepackage{array}
\usepackage[caption=false,font=normalsize,labelfont=sf,textfont=sf]{subfig}
\usepackage{textcomp}
\usepackage{stfloats}
\usepackage{url}
\usepackage{verbatim}
\usepackage{graphicx}
\usepackage{cite}

\usepackage{xcolor}
\usepackage[dvipsnames]{xcolor}

\usepackage{overpic}     
\usepackage{multirow}   
\usepackage{multicol}    
\usepackage{colortbl}    
\usepackage{booktabs}    
\usepackage[colorlinks=true,linkcolor=blue,citecolor=blue,urlcolor=blue]{hyperref}
\usepackage[capitalize]{cleveref}
\usepackage{tikz,xcolor,hyperref}% Make Orcid icon
\definecolor{lime}{HTML}{A6CE39}
\DeclareRobustCommand{\orcidicon}{%
    \begin{tikzpicture}
    \draw[lime, fill=lime] (0,0) 
    circle [radius=0.16] 
    node[white] {{\fontfamily{qag}\selectfont \tiny ID}};    \draw[white, fill=white] (-0.0625,0.095) 
    circle [radius=0.007];    \end{tikzpicture}
    \hspace{-2mm}}
\foreach \x in {A, ..., Z}{%
    \expandafter\xdef\csname orcid\x\endcsname{\noexpand\href{https://orcid.org/\csname orcidauthor\x\endcsname}{\noexpand\orcidicon}}
    }
\makeatletter
\def\onedot{\ifx\@let@token.\else.\null\fi\xspace}
\makeatother

\newcommand{\methodname}{RAM++}

\def\eg{\emph{e.g}\onedot} 

\def\ie{\emph{i.e}\onedot}

\newcommand{\secref}[1]{Sec.~\ref{#1}}

\definecolor{picred}{RGB}{153,34,46}
\begin{document}
% 放在导言区（\documentclass 后）
\setlength{\abovecaptionskip}{-5pt}  % caption 到“正文/图”的上方间距
\setlength{\belowcaptionskip}{0pt}  % caption 到下方内容的间距

\bstctlcite{BSTcontrol}

\title{RAM++: \underline{R}obust Representation Learning via \underline{A}daptive \underline{M}ask for All-in-One Image Restoration}

\author{Zilong Zhang\orcidA{}, Chujie Qin\orcidB{}, Chunle Guo\orcidC{}, Yong Zhang, Chao Xue,\\ Ming-Ming Cheng\orcidF{},~\IEEEmembership{Senior~Member,~IEEE} and Chongyi Li\orcidG{},~\IEEEmembership{Senior~Member,~IEEE}
\thanks{This work was supported in part by the Natural Science Foundation of Tianjin, China (24JCJQJC00020),  in part by 
the Fundamental Research Funds for the Central Universities (Nankai University, 070-63243143).
An earlier version of this paper was presented at ECCV 2024 (DOI: 10.1007/978-3-031-72995-9\_21).
\textit{(Z.~Zhang and C.~Qin contributed equally to this work.) (Corresponding author: C.~Li.)}

Z. Zhang, C. Qin, C. Guo, M.-M. Cheng, and C. Li are with VCIP, CS, Nankai University, Tianjin 300350, China. (e-mail: \{zhangzilong, chujie.qin\}@mail.nankai.edu.cn; \{guochunle, cmm, lichongyi\}@nankai.edu.cn).

Y. Zhang is with Chongqing Chang'an Wangjiang Industrial Group Co., Ltd and also with VCIP, CS, Nankai University. (e-mail: zhangyongtju@163.com).

C. Xue is with Tiandy Technologies. (e-mail: xuechao@tiandy.com).%
}}

%\thanks{This paper was produced by the IEEE Publication Technology Group. They are in Piscataway, NJ.}% <-this % stops a space


% % column for its text to clear the IEEEpubid mark.

\maketitle

\begin{abstract}
This work presents \underline{R}obust Representation Learning via \underline{A}daptive \underline{M}ask (\methodname), a two-stage framework for all-in-one image restoration. \methodname{} integrates high-level semantic understanding with low-level texture generation to achieve content-oriented robust restoration. It addresses the limitations of existing degradation-oriented methods in extreme scenarios (\eg, degradations are strongly coupled with image structures).  \methodname{} also mitigates common challenges such as unbalanced performance across tasks, overfitting to seen degradations, and weak generalization to unseen ones through three key designs: 1) Adaptive Semantic-Aware Mask (AdaSAM): a pre-training strategy that applies pixel-level masks to semantically rich and textured regions. This design enables the network to learn both generative priors and image content priors from various degradations. 
2) Mask Attribute Conductance (MAC): a selective fine-tuning strategy that adjusts the layers with higher contributions to bridge the integrity gap between masked pre-training and full-image fine-tuning, while retaining learned priors.
3) Robust Feature Regularization (RFR): a strategy that leverages DINOv2’s semantically consistent and degradation-invariant representations, together with efficient feature fusion, to achieve faithful and semantically coherent restoration.
With these designs,  \methodname{} achieves robust, well-balanced, and state-of-the-art performance across seen, unseen, extreme, and mixed degradations.
Our code and model will be released at \url{https://github.com/DragonisCV/RAM}

\end{abstract}


\begin{figure*}[t]
  \centering
   \includegraphics[width=0.95\linewidth]{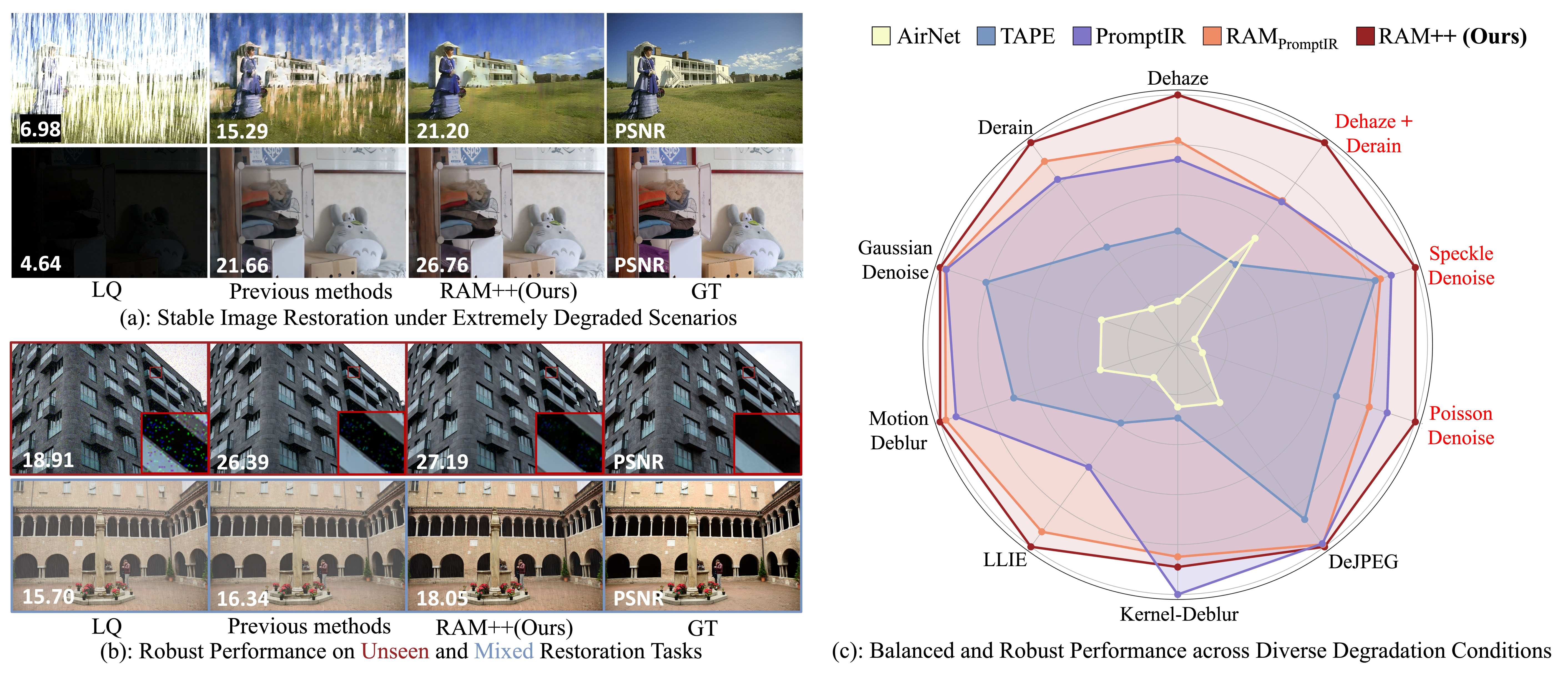}
   \vspace{2pt}
   \caption{\text{RAM++} outperforms state-of-the-art methods (\eg, PromptIR~\cite{promptir}, HOGFormer~\cite{hogformer}) in all-in-one blind image restoration, delivering more balanced and robust performance across seen, extreme,  and \textcolor{red}{unseen} (single and mixed) degradations.}
   \label{fig:performance}
\end{figure*}

\section{Introduction}
\label{sec:intro}

\IEEEPARstart{I}{mage} restoration refers to the process of recovering high-quality images from low-quality ones affected by various types of degradation, typically caused by adverse environmental conditions (\eg, rain, haze, insufficient illumination), hardware limitations (\eg, noise and blur), or post-processing artifacts (\eg, JPEG compression).
Image restoration not only improves perceptual quality but also supports practical applications such as autonomous driving and video surveillance.

Recent techniques in this field focus mainly on learning fixed degradation patterns, \ie degradation priors.
Some approaches~\cite{rain_1, lowlight2,li2016underwater} utilize task-specific priors to address a certain degradation type, while another line of work~\cite{liang2021swinir,Zamir2021Restormer,mehri2021mprnet,nafnet} attempts to design general network architectures that can effectively capture different degradation patterns.
However, these methods are usually trained to handle only one type of degradation at a time. As a result, their performance becomes imbalanced when faced with multiple types of degradation.

To address this problem, all-in-one methods have been proposed to restore multiple degradations using a single model.
Most of these methods tend to incorporate explicit priors (\eg, AirNet~\cite{airnet}), add auxiliary modules (\eg, PromptIR~\cite{promptir}), or introduce pre-training with degradation classification (\eg, DCPT~\cite{hu2025universalimagerestorationpretraining}) to identify degradation types and assist restoration.
However, these approaches focus on distinguishing degradation types in an image rather than modeling its actual content. Thus, they often suffer from the following limitations.

\textbf{Insufficient generative capability.}
In the deep learning era, image restoration is essentially generating high-quality images from low-quality inputs. Conventional all-in-one restoration models~\cite{airnet,moceir,instructir}, trained on paired degraded–clean images, typically learn only a direct degradation-to-clean mapping, failing to capture latent structures or acquire generative capabilities. Under extreme degradations, as they cannot infer missing details, these models often struggle to restore severely degraded regions, as shown in Fig.~\ref{fig:performance}(a). In fact, when pixel information is severely lost, these models fail to reconstruct plausible structures, resulting in frustrating outcomes that are intolerable for high-quality restoration.
% In the deep learning era, image restoration can be viewed as generating high-quality images conditioned on low-quality inputs. However, conventional all-in-one image restoration models~\cite{airnet,moceir,instructir}, trained on paired degraded–clean images, typically only learn the direct mapping from degradation to clean images and fail to capture latent image structures or acquire generative capabilities. Specifically, when trained under diverse degradation types, such models often fail to achieve thorough restoration under extreme degradations, particularly in severely degraded regions (see Fig.~\ref{fig:performance}a). This failure stems from the models’ inability to infer missing or heavily degraded details. When pixel information is nearly lost, they cannot “imagine” or generate plausible structures, resulting in poor restoration. Such failures are unacceptable for a high-quality restoration model.

\textbf{Unbalanced task performance.}
Mixed training scenarios inherently require shared representations across degradation types. Nevertheless, degradation-oriented models~\cite{airnet,promptir} tend to map images of different degradation types into distinct feature spaces (as evidenced by the t-SNE distribution in Fig.~\ref{fig:tsne_comparison}). This leads to substantial discrepancies among their representations and even causes trade-offs between tasks. Consequently, performance becomes unbalanced across degradation types, with some degradation types experiencing severe collapse.

\textbf{Limited generalization ability.}
Due to the degradation-classification-oriented learning paradigm adopted by prior models~\cite{hu2025universalimagerestorationpretraining}, as well as the inherent limitations of architectures with long-range residual connections (see \secref{sec:mim_discussion}), these models tend to overfit degradation patterns severely. Thus, they struggle to generalize to unseen single or mixed degradation scenarios (Fig.~\ref{fig:performance}(b)). This limitation not only hinders practical deployment but also fundamentally deviates from the original intent of the all-in-one image restoration task.

% which limits their scalability and leads to ambiguous decision boundaries as the number of degradation types increases.
% We argue that the essence of image restoration lies in learning a robust intrinsic representation that recovers the essential image information from degraded inputs, rather than merely removing degradation patterns—\ie, focusing on image priors instead of degradation priors.
% TAPE~\cite{liu2022tape} also suggests that modeling the natural properties of clean images can benefit restoration by introducing a natural image prior.
% However, TAPE uses the network’s own output as the supervision target, which can cause the model to reinforce its own mistakes and learn a biased image prior.
\begin{figure}[h!t]
\centering
\includegraphics[width=0.95\linewidth]{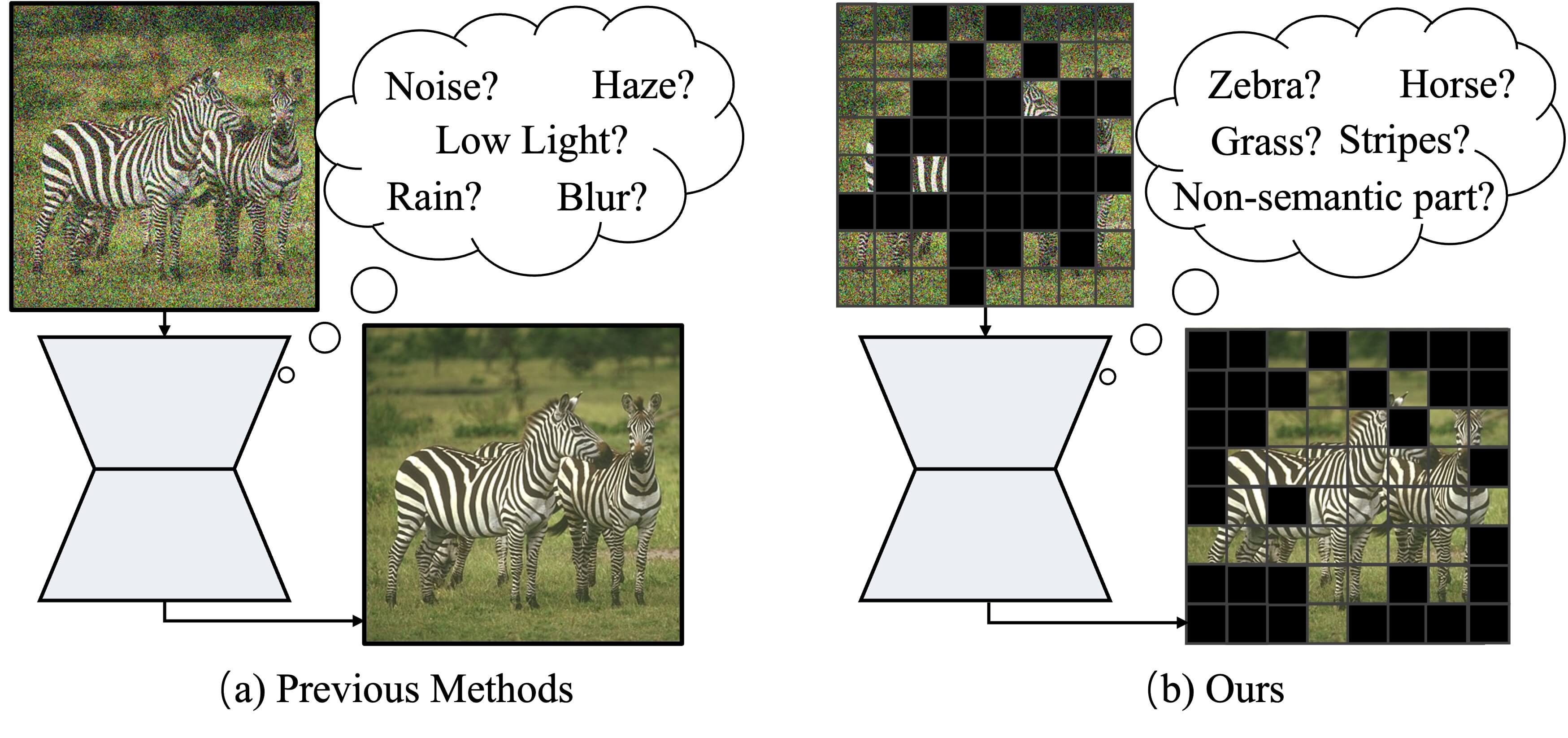}
\vspace{2pt}
\caption{A high-level view of our method compared with previous methods. Our approach emphasizes image content priors rather than degradation distinctions.}
\label{fig:teaser}
\end{figure}

Considering the above problems, we argue that the essence of image restoration lies in learning a robust intrinsic representation that recovers the essential image information from degraded inputs, rather than merely removing degradation patterns, \ie, focusing on image priors instead of degradation priors, as illustrated in Fig.~\ref{fig:teaser}. 
In this work, we aim to solve the problem of \textit{how to extract robust intrinsic image information from various degraded images}.

Recent efforts~\cite{mim_lowlevel,mim_denoising} using Masked Image Modeling (MIM)~\cite{mae,simmim} in low-level vision have attracted our attention.
As a pre-training strategy, MIM has been widely verified to be effective for high-level vision tasks, owing to its ability to learn general image representations.
At the same time, it encourages the model to learn the distribution of natural images, which includes the intrinsic information we aim to seek.
Based on MIM, we propose a simple framework, \textbf{\methodname}, for all-in-one blind image restoration to learn a robust representation via adaptive masks. 
Our method consists of two stages: an Adaptive Semantic-Aware Mask (AdaSAM) based pre-training stage and a fine-tuning stage guided by Mask Attribute Conductance (MAC) and Robust Feature Regularization (RFR), leveraging a vision foundation model (\eg, DINOv2~\cite{dinov2}) with degradation-invariant priors.

In the pre-training stage, we train the AdaSAM network to selectively mask pixels in degraded images, focusing on richly textured regions around the main subjects, which are inherently more difficult to restore. The restoration network is then tasked with reconstructing the clean content at these masked locations, thereby encouraging it to learn intrinsic image priors and to encode diverse degradations into a more unified and semantically consistent feature space.

In the fine-tuning stage, we leverage DINOv2's robust representations to handle various degraded images, overcoming the input inconsistency caused by the difference between masked inputs used during pre-training and the full-image input during inference. At the same time, we use MAC to preserve the learned prior as much as possible.
Specifically, we first evaluate the contribution of each network layer in addressing this inconsistency using the proposed MAC.
Then, we select the top $k\%$ most important layers for fine-tuning while freezing the remaining ones.
We show that even with a short fine-tuning phase (\eg, updating only $30\%$ of layers), the model achieves strong performance and outperforms models trained using traditional pair-wise supervision.

Additionally, we observe that DINOv2 layer-wise features exhibit degradation-independent semantic representations. To fully leverage this capability, we develop an efficient module for feature modulation and fusion. 
% we develop an efficient DINOv2 feature modulation module and feature fusion module. 
% This 
The fused features enhance restoration performance by facilitating essential information extraction from images.
Fig.~\ref{fig:tsne_comparison} demonstrates that the proposed pre-training strategy effectively encodes various degraded images into a unified feature space, indicating improved robustness and semantic consistency in the learned representations. Furthermore, after fine-tuning, the models show an inherent capability to classify degradations, even without any supervision for degradation classification. Fig.~\ref{fig:performance}(c) shows that \methodname{} outperforms all state-of-the-art all-in-one approaches and demonstrates strong generalization in complex scenarios.
% Moreover, our pipeline can be directly applied to existing networks without introducing any additional computational cost.

This work builds upon our earlier conference version~\cite{qin2024restore}. Compared to the previous version, we introduce several significant extensions and improvements, as detailed below:
\begin{figure*}
    \centering
    \includegraphics[width=0.95\linewidth]{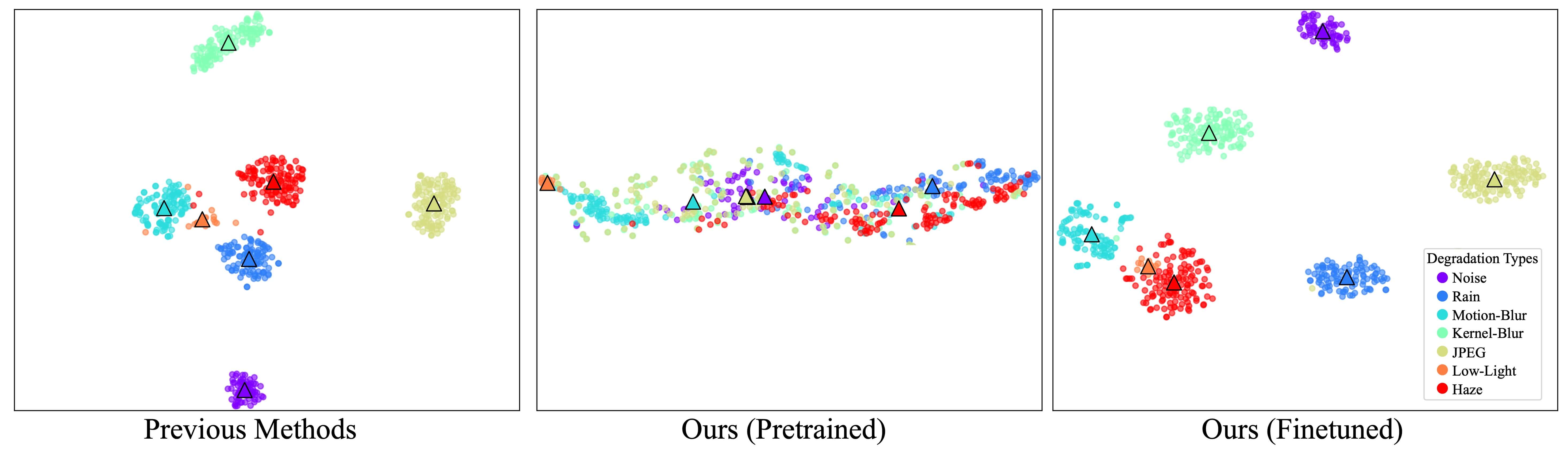}
    \vspace{2pt}
    \caption{T-SNE visualization of features from all-in-one trained \textit{Previous Methods}(\eg, Restormer\cite{Zamir2021Restormer})   (left), the features pre-trained with \textit{Adaptive Semantic-Aware Mask} (center), and the features after fine-tuning with the \textit{DINOv2 regularization} (right). This visualization demonstrates the model’s delicate balance between shared feature representation and degradation-specific discrimination ability.
    % task-specific requirements.
    }
    \label{fig:tsne_comparison}
\end{figure*}
\begin{itemize}
\item We identify the challenges of applying MIM in low-level vision and propose an Adaptive Semantic-Aware Mask Pre-training strategy based on MIM for all-in-one blind image restoration. Unlike the conference version~\cite{qin2024restore}, which uses overly simple random pixel-level masks easily recovered by interpolation, our mask strategy integrates high-level semantics with low-level textures, enabling the model to concentrate on challenging-to-reconstruct regions and to aggregate diverse degradations into a more unified latent space, facilitating the learning of more robust and semantic-aware representations.
%\item We design an A to guide the network's focus toward challenging regions by computing the semantic correlation across different regions of the input image, encouraging the model .
% \item We propose Mask Attribute Conductance to assess the importance of each layer in bridging the input inconsistency, so that only a small portion (\eg, $30\%$) of critical layers are updated, preserving the learned image prior from MIM.
\item We propose Robust Feature Regularization, which fuses model features with a pretrained representation extractor during fine-tuning. This approach stabilizes feature representations, enhances generalization across diverse degradations, and ensures reliable performance under challenging conditions, further validating the effectiveness of focusing on intrinsic image information.
\item In comparison to the conference version~\cite{qin2024restore}, we further conduct extensive experiments under three different task-scale settings, compare our method with the latest state-of-the-art approaches, and discuss emerging related methods. 
We also perform more rigorous evaluations on a broader range of standard benchmarks in this field. 
Experimental results demonstrate that \methodname{} achieves the best and most balanced performance on seen degradations, while showing substantial improvements on unseen degradations, highlighting the broad applicability of the proposed framework.
\item We further adopt the latest interpretability techniques (\eg, CEM~\cite{cem}) for low-level vision to investigate the underlying causes of the performance improvements. 
Analysis reveals that \methodname{} possesses four remarkable properties: effective semantic understanding, stable global information acquisition, accurate positive/negative information discrimination, and prioritized background structure reconstruction. 
Overall, this work offers a novel content-oriented perspective for all-in-one image restoration and achieves significant results.
% \item We conducted extensive experiments under 3-, 5-, and 7-task settings, and evaluated on nearly all standard benchmarks in this field. These results show that the proposed \methodname{} method achieves state-of-the-art and well-balanced performance on seen degradations, and demonstrates substantial improvement on unseen degradations, offering a new perspective for all-in-one image restoration.
% \item This extended version not only surpasses the conference version~\cite{qin2024restore} in terms of performance and robustness, but also employ the latest interpretability techniques in low-level vision to analyze the underlying causes of performance gains. It further provides a more rigorous evaluation by incorporating comparisons with recent state-of-the-art methods, discussing emerging related approaches, and examining adaptability under varying task numbers. Particular emphasis is placed on generalization across diverse restoration scenarios, underscoring the broader applicability of the proposed framework.

% \item Our proposed RAM++ provides a new perspective for achieving more balanced and effective all-in-one blind image restoration by focusing on recovering intrinsic image information from degraded images.
% The framework is compatible with any image restoration backbone and does not introduce additional computational overhead.
\end{itemize}
\section{Related Work}
\label{sec:related_work}

\subsection{Image Restoration for Multi-Degradations}
Although neural networks have achieved remarkable results in single degradation image restoration~\cite{deblur_1,guo2020zerodce,guo2022dehamer,wu2023ridcp,lowlight2,rain_1,jin2023dnf,li2021underwater,li2023embedding}, recent studies have gradually shifted their attention towards the more challenging task of handling multiple degradations simultaneously.
A class of methods~\cite{nafnet,Zamir2021Restormer,liang2021swinir,mehri2021mprnet} focuses on designing general architectures capable of effectively capturing different degradation patterns.
SwinIR~\cite{liang2021swinir} applies a window-based attention mechanism to localize global attention, thereby reducing computational cost.
Similarly, transformer-based U-shaped models~\cite{wang2022uformer,Zamir2021Restormer} are adopted to extract multi-scale features while maintaining computational efficiency.
However, these methods typically require separate training for each restoration task.
Some other approaches~\cite{li2020all} utilize multiple input and output heads to enable the network to handle various degradation types.
Nevertheless, such designs often result in limited scalability.
Recently, several methods~\cite{airnet,promptir,expert,hu2025universalimagerestorationpretraining} have been proposed to address multiple restoration tasks using a unified model.
Most of them focus on learning to differentiate between degradation types and restoring degraded content accordingly.
For instance, AirNet~\cite{airnet} is among the first to propose an all-in-one image restoration framework.
It begins by pre-training a degradation classifier using contrastive learning and then uses it to guide the restoration process.
PromptIR~\cite{promptir} introduces a learnable, prompt-based module that, rather than relying on fixed degradation categories, enables the model to autonomously learn features that enhance its performance through adaptive prompting.
To improve the model’s ability to distinguish degradations in low-quality inputs, DCPT~\cite{hu2025universalimagerestorationpretraining} introduces a pre-training stage that exposes the model to diverse degradations early on, allowing it to learn degradation-aware representations before full restoration training.
In contrast, our proposed \methodname{} offers a new perspective by focusing on extracting shared content information from degraded images, without incorporating any additional components for degradation recognition.
This design enables us to achieve both balanced and strong performance, especially when addressing a wider range of degradation types.
%----------------------------------------------------------------------
\subsection{Mask Image Modeling}
Inspired by Masked Language Modeling~\cite{bert}, MIM~\cite{mae,simmim} has been proposed as a pre-training strategy to learn general-purpose representations in high-level vision tasks.
MAE~\cite{mae} leverages MIM by predicting missing tokens, showing strong performance and good generalization across a wide range of downstream applications.
SimMIM~\cite{simmim} presents a generic MIM approach based on Swin-ViT~\cite{swin}.
Painter~\cite{painter} unifies various tasks into image-to-image translation and benefits from MIM-based pre-training.
AdaMAE~\cite{Bandara2022AdaMAEAM} trains the sampling network by masking each frame of the video with a high masking ratio, preserving high-information regions.
% , in contrast to our approach.
%
Some attempts have been made to extend MIM to low-level vision, aiming to improve model generalization.
Among them, \cite{mim_lowlevel} and~\cite{mim_denoising} are most relevant to our work.
\cite{mim_denoising} adopts MIM to enhance generalization in denoising tasks but does not explore its application in multi-task settings.
\cite{mim_lowlevel} uses MIM to pre-train an encoder for introducing generative priors and then uses a decoder for restoration, but it does not fully utilize the potential of MIM.
Our proposed \methodname{} applies MIM to unify the learning objective across different restoration tasks into reconstructing intrinsic image information.
This design helps the network learn restoration functions in a more balanced and effective manner.
To further retain the image priors acquired during MIM pre-training, we develop a fine-tuning strategy based on MAC analysis (as described in~\secref{sec:finetuning_stage}).
This allows the restoration model to achieve competitive performance by updating only a small fraction (\eg, $30\%$) of layers, maximizing the advantages from MIM.

%----------------------------------------------------------------------

\subsection{Gradient-based Attribution}
Gradient-based attribution methods~\cite{ig,approx_ig,layerconductance,lam,faig} are widely adopted to interpret how hidden units (or inputs) influence the network’s output.
A commonly used technique is Integrated Gradients (IG)~\cite{ig,approx_ig}, which computes the accumulated gradients along a linear path from a baseline input to the actual input in the pixel or feature space.
Subsequently, methods such as IntInf~\cite{intinf} and layer conductance~\cite{layerconductance} extend IG to assess the importance of neurons along the same path.
In our work, we aim to identify the key layers that are most effective in bridging the distribution gap between training and inference data.
We propose Mask Attribute Conductance, which builds upon layer conductance by accumulating contributions from each layer along the Mask Attribute Path.
Mask Attribute Conductance serves as a measure of each layer’s importance along this path.
Based on the Mask Attribute Conductance values, we selectively fine-tune the top $k\%$ most critical layers of the pre-trained network, thereby preserving the image priors learned during the pretraining stage to a large extent.

%----------------------------------------------------------------------
\subsection{Pre-trained Large-scale Vision Foundation Models}
Vision foundation models~\cite{CLIP,dinov2} are typically trained on millions of images, providing representations with strong generalization and transfer capabilities. 
Therefore, these pre-trained foundation models can be fully applied to downstream tasks with only simple natural language descriptions or prompts. 
In recent years, several works have introduced pre-trained models into the field of image restoration.
CLIP-AWR~\cite{tan2024exploring} integrates CLIP-driven spatial feature extraction with semantic prior embedding to address diverse adverse weather scenarios in image restoration.
Similarly, DA-CLIP~\cite{luo2023controlling} introduces an auxiliary controller that refines the frozen CLIP image encoder, producing more expressive feature embeddings to guide high-fidelity reconstruction.
DINO-IR~\cite{lin2023multi} dynamically fuses shallow pixel-level features and deep degradation-invariant semantic features extracted from DINOv2. 
Perceive-IR~\cite{zhang2025perceive} introduces a prompt learning mechanism that constrains the similarity between prompts and images in the CLIP perception space.  
Our proposed Robust Feature Regularization leverages DINOv2 to emphasize consistency across adjacent hierarchical representations and robustness to diverse degradations. Simultaneously, the rich semantic features of DINO help mitigate discrepancies in input completeness during the two-stage training process.

\section{Methodology}
In this section, we start by discussing the challenges of using MIM in low-level vision tasks (\secref{sec:mim_discussion}). 
Following that, we present our pipeline for all-in-one blind image restoration, which contains three parts: pre-training with MIM (\secref{sec:pretraining_stage}), fine-tuning with Mask Attribute Conductance (MAC) Analysis (\secref{sec:finetuning_stage}), and feature restoration assisted by DINOv2 (\secref{sec:dino_feature}). Fig.~\ref{fig:overall pipeline} illustrates the overall pipeline.

\subsection{Rethinking MIM in Low-Level Vision}
\label{sec:mim_discussion}
\begin{figure}[h]
    \centering
    \includegraphics[width=\linewidth]
    {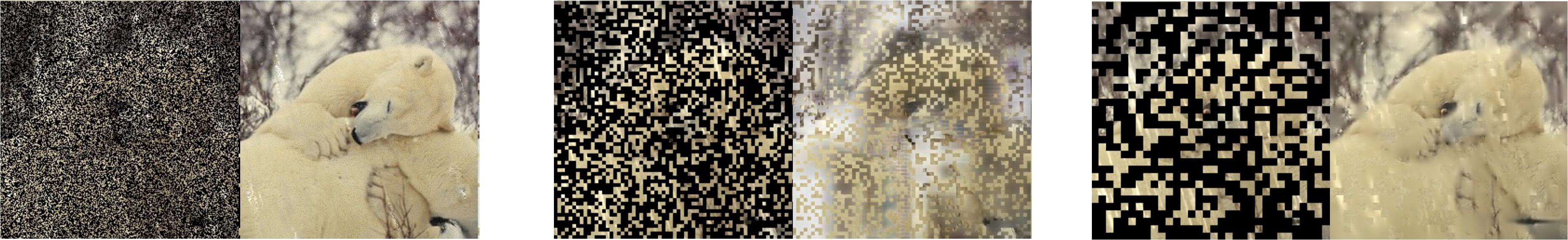}
    \put(-237,-10){patch size=1}
    \put(-149,-10){patch size=4}
    \put(-64,-10){patch size=8}
    \vspace{10pt}
    \caption{Mask Image Modeling reconstruction with different patch sizes. We pre-trained with different patch sizes and visualized the mask inputs (left), and the corresponding MIM reconstructions (right). This suggests that 
    \textit{pixel-level} masks are necessary for low-level image restoration tasks.}
    \label{fig:patch-size}
\end{figure}

MIM is a process that randomly masks parts of an image and extracts features from the remaining visible regions to reconstruct the complete image.
This enables models to learn a general representation of images, leading to effective pre-training, which has been validated across various high-level tasks~\cite{mae,simmim}.
In addition, the model also captures the distribution of natural images during the reconstruction process, \ie, through MIM pre-training.
Such incidental learning of prior knowledge proves beneficial for tasks like image restoration.
However, despite these strengths, applying MIM as a pre-training strategy for low-level vision tasks remains under-explored, mainly due to the unique challenges as follows:

\begin{figure*}[!t]
  \centering
   \includegraphics[width=0.95\linewidth]{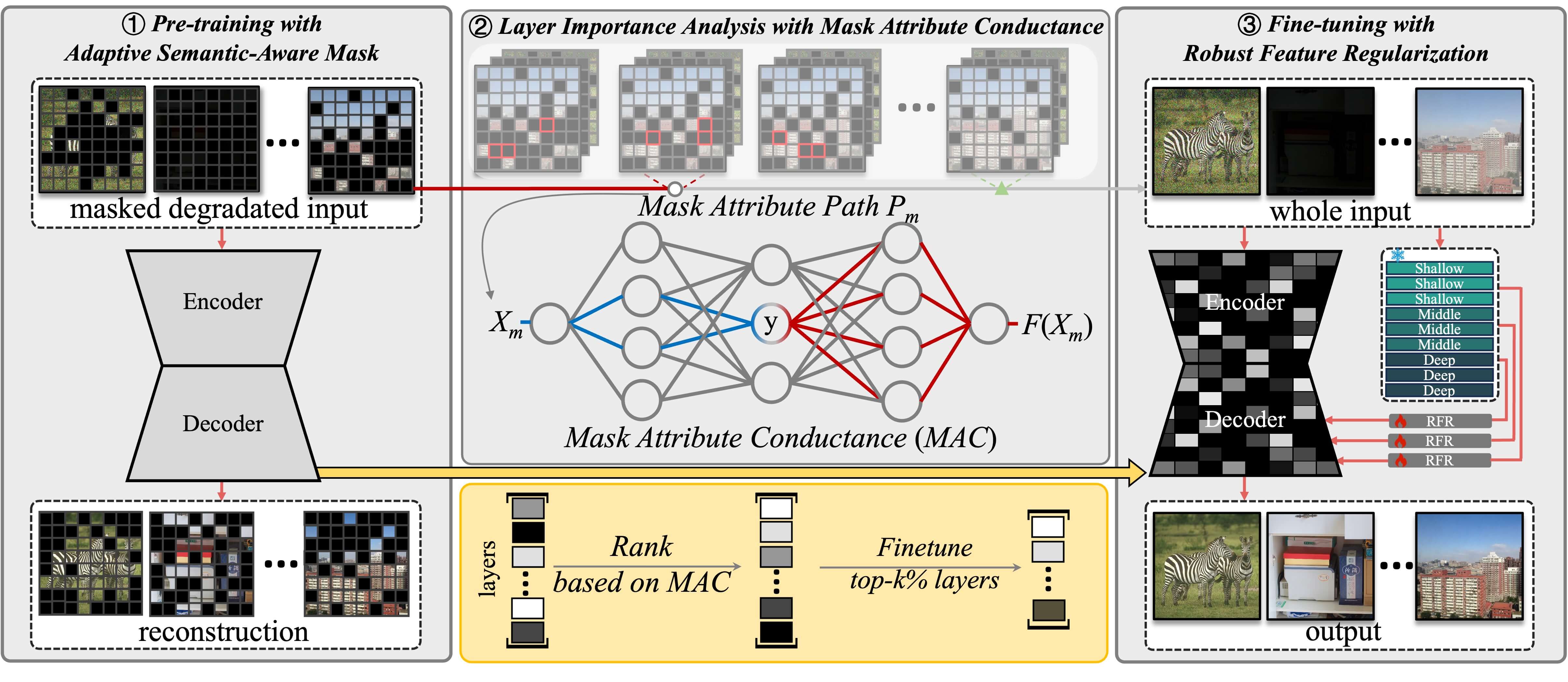}
   \caption{An illustration of our overall pipeline. 1) Pre-training the model with the Adaptive Semantic-Aware Mask image method tailored to low-level vision. We mask the degraded images’ semantically and texturally rich regions (\ie, high-information regions) at the pixel level with a $50\%$ masking ratio and reconstruct the clean images.
   2) Fine-tuning is performed to bridge the input integrity gap that arises when transitioning from masked inputs during pre-training to full images during inference. We assess the contribution of each network layer to addressing this gap using the proposed MAC, ranking them in descending order. The top $k\%$ of layers are then selected for fine-tuning on complete images.
   3) The fine-tuning process is further assisted by a pre-trained vision foundation model, providing semantic consistency and degradation-invariant priors.}
   \label{fig:overall pipeline}
   \vspace{-4mm}
\end{figure*}
1) The original purpose of vanilla MIM is not to achieve high-quality reconstruction, but to learn useful features for high-level vision tasks.
As a result, it tends to mask a larger portion of the image to focus on semantic understanding rather than pixel-level detail, which is reflected in token-level masking and a high masking ratio.
CSFormer~\cite{mim_lowlevel} directly adopts this strategy for low-level vision pre-training.
However, some studies have shown that, unlike in pattern recognition tasks, relying solely on semantic information while neglecting texture details is suboptimal for image restoration~\cite{liu2021discovering, magid2022texture}.
% However, some studies have shown that semantic information is less critical for image restoration than it is for recognition tasks~\cite{liu2021discovering, magid2022texture}.
In addition, using a high-degree masking can lead to loss of fine details in the reconstructed output, as shown in Fig.~\ref{fig:patch-size}, which negatively affects performance in low-level tasks.
\begin{figure}[h]
  \centering
   \includegraphics[width=0.95\linewidth]{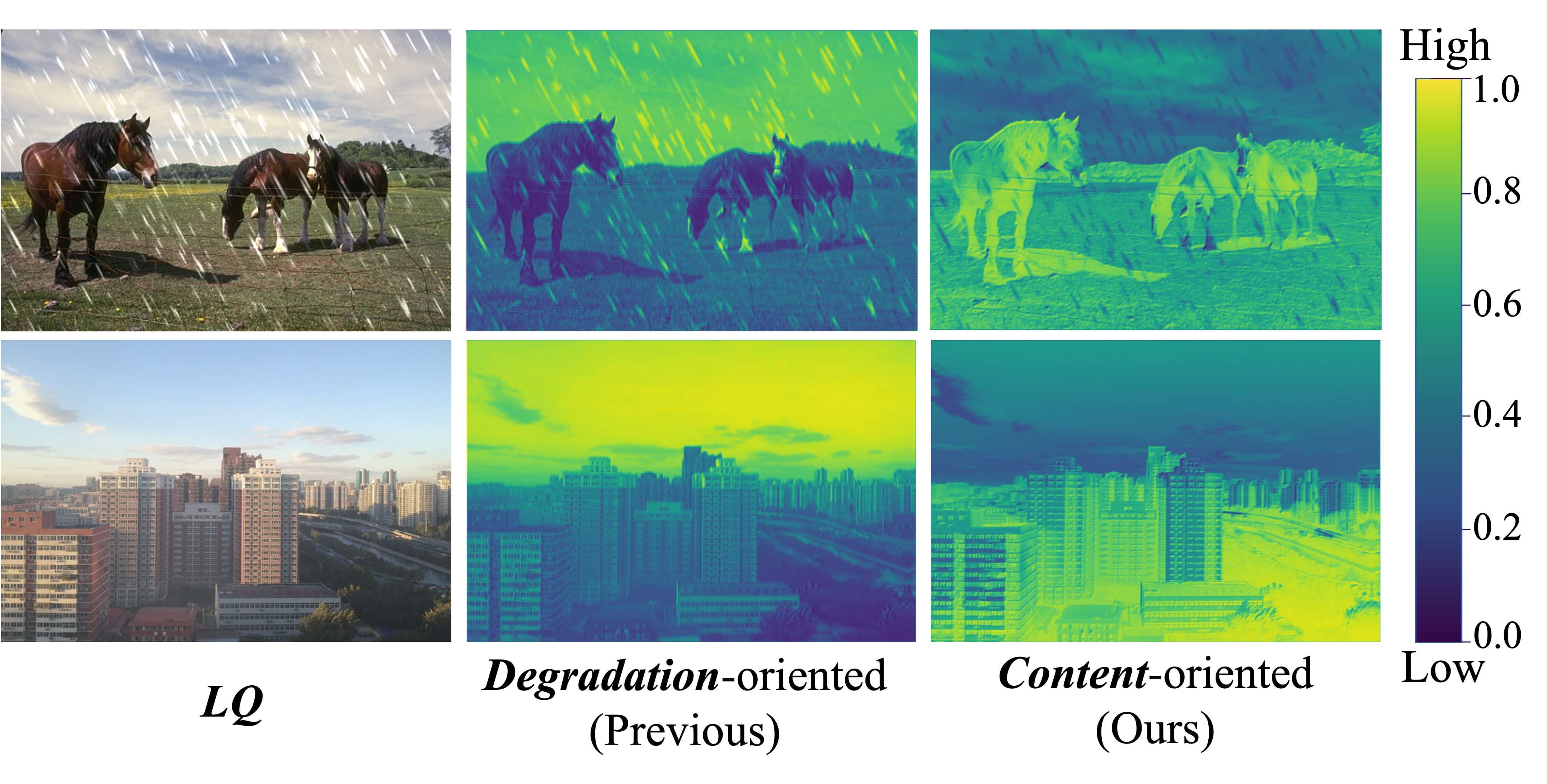}
   \caption{Visualization of feature maps from our content-oriented restoration model and previous degradation-oriented models~\cite{promptir}. The comparison reveals that our model emphasizes structural details and semantic consistency, rather than the differences among degradations.}
   \label{fig:content_oriented}
\end{figure}

2) The objective of MIM is to reconstruct the masked version of the input image, so the outputs remain in the same domain as the inputs.
However, in image restoration, we expect the model to bridge the domain gap between degraded inputs and clean outputs, \ie, recover high-quality content from low-quality inputs.
To achieve this, it's necessary to use paired data during MIM pre-training for restoration tasks (see the experiment in~\secref{sec:ablation} for details).~\cite{mim_denoising} show that pair-wise MIM training improves generalization across different types of noise.
In this paper, we further investigate the effectiveness of MIM for handling multiple degradations with greater variation.

3) Mainstream image restoration models typically employ a global residual connection, formulated as $\hat{y} = M(x) + x$, where $\hat{y}$ denotes the predicted high-quality image, $x$ is the low-quality input, and $M$ represents the restoration model~\cite{jiang2024survey}. This design inherently drives the model to focus on identifying degradation features and correcting degraded regions, rather than reconstructing masked areas or capturing the essential semantics across different images~\cite{liu2021discovering}. As such, this paradigm is not well aligned with the generalization objective of our MIM framework. Instead, we tentatively rely solely on the internal residual structures within the model, \ie, modeling it as $\hat{y} = M(x)$, aiming to learn meaningful content representations while preserving as much fine-grained image detail as possible. Fig.~\ref{fig:content_oriented} illustrates that degradation-driven methods primarily produce feature maps responding to low-level degradations, whereas our content-oriented approach captures more semantically meaningful and structurally coherent representations.
 Although no explicit constraint is imposed on degradation discrimination, our model spontaneously learns to distinguish various types of degraded images and exhibits the ability to recognize unknown degradations after fine-tuning.
 \begin{figure}[t]
    \centering
    \includegraphics[width=0.95\linewidth]{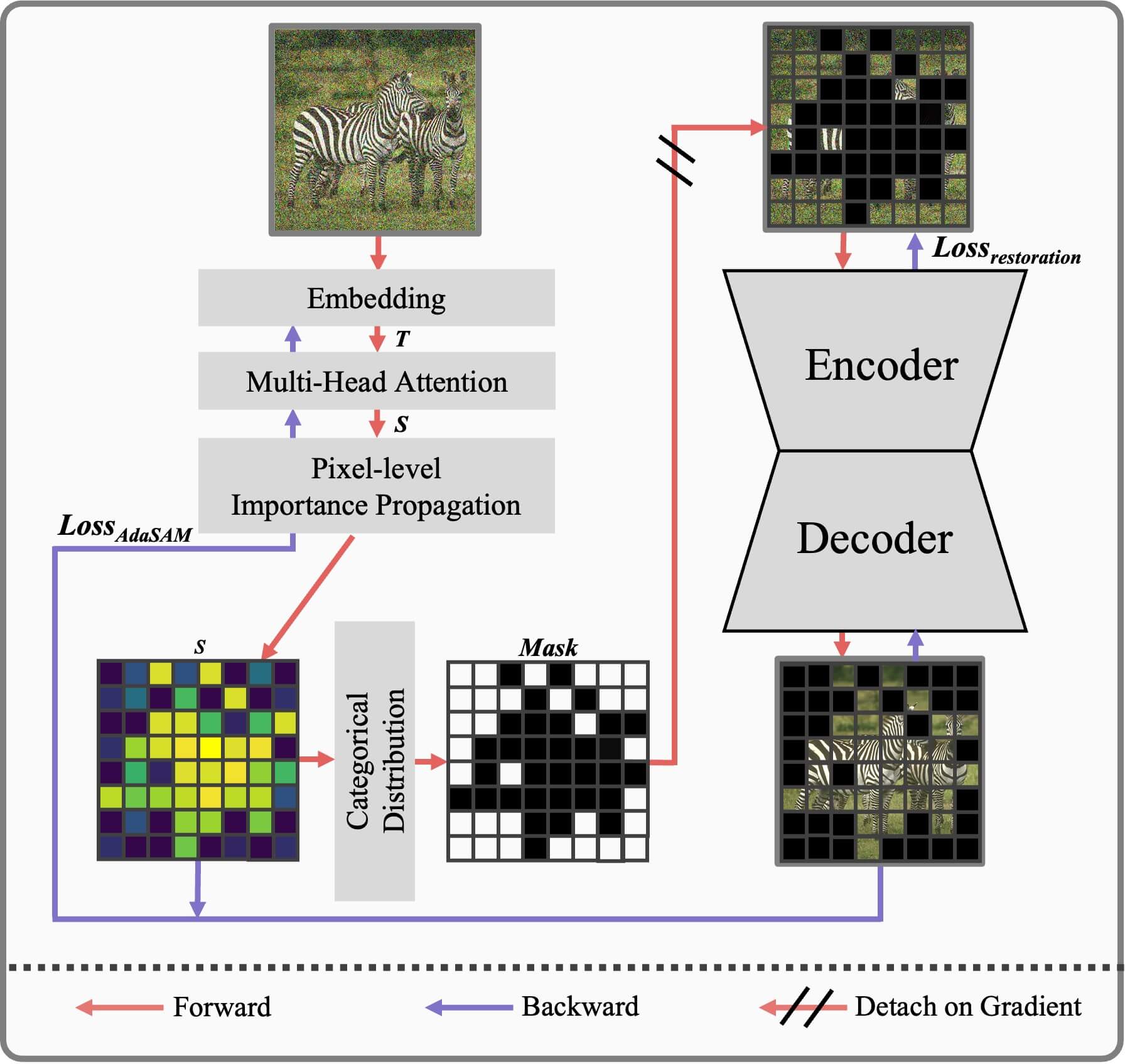}
    \caption{Overall architecture of AdaSAM: masks from high-information regions guide the model to reconstruct structures from low-information areas.}
    \label{fig:adasam}
\end{figure}
\subsection{Adaptive Semantic-Aware Mask Pre-training.}
\label{sec:pretraining_stage}
Based on the above analysis, we design a MIM pre-training paradigm tailored for low-level vision.
To prevent the model from trivially minimizing loss through simple interpolation under random pixel-level masking, we combine pixel-level restoration with region-level semantic understanding. This design encourages the model to ignore degradation types and instead focus on reconstructing intrinsic image content.
To this end, we propose an Adaptive Semantic-Aware Mask (AdaSAM) strategy. This masking approach generates adaptive masks for the fully degraded input images by estimating the importance of each region based on semantic relevance.
In the following, we present the design of AdaSAM in detail, including its core components and the corresponding optimization objectives.
% \paragraph{Masking.} During the pre-training stage, we randomly mask the pixels of degraded images (mask images in a $1 \times 1$ patch size) with a $50\%$ mask ratio.
% We found that fine-grained masked patches and balanced mask ratio are beneficial to image restoration, which can be demonstrated in Sec.~\ref{sec:ablation}.

% Besides, since our MIM pre-training has a similar target to subsequent low-level tasks, we do not need to change the decoder like MAE~\cite{mae} does but just fine-tune it.
\paragraph{Core Components}
% To encourage the model to ignore distinctions among different degradations while creating a more challenging restoration task, we integrate pixel-level restoration with region-level semantic understanding.
%
%
As illustrated in Fig.~\ref{fig:adasam}, our masking pipeline comprises four components: Patch Embedding, Multi-Head Attention (Atten), Pixel-level 
Importance Propagation, and Categorical Sampling. The first two modules collaborate to compute region-level semantic importance scores, while the latter two generate the pixel-level mask based on these scores.

Specifically, given a degraded image \( I_d \in \mathbb{R}^{C \times H \times W} \), we first encode it into 
$N = \frac{H}{h} \times \frac{W}{w}$,
non-overlapping tokens using a patch embedding module. These tokens \( T \) serve as the fundamental feature representations for subsequent processing. Next, the token sequence \( T \) is fed into the Attention module, which captures long-range contextual semantics across spatial regions. The output of the Attention module is then passed through a linear projection layer followed by softmax normalization, resulting in token-wise importance scores: 
% These tokens  are then processed by the Attention module to capture contextual semantics, followed by a linear projection and softmax activation to obtain token-wise importance scores:
\begin{equation}
    S = \mathrm{Softmax}(\mathrm{Linear}(\mathrm{Atten}(T))).
\end{equation}
This score essentially reflects the relational dependency of a token with respect to the others.
Rather than applying coarse-grained masking over large patches~\cite{mae,Bandara2022AdaMAEAM}, 
we uniformly propagate each token’s importance score to all pixels within its corresponding spatial 
region, thereby constructing a fine-grained, pixel-level importance map $s$ across the entire image. 
This map defines a Categorical Distribution, in which pixels belonging to more informative tokens 
are assigned higher selection probabilities.

Given a predefined mask ratio $\rho$, the exact number of masked pixels is computed as
\begin{equation}
    N_{\text{mask}} = H \times W \times  \rho.
\end{equation}
We perform multinomial sampling without replacement to select candidate pixels:
\begin{equation}
    \mathcal{I}_{\text{mask}} \sim \text{Multinomial}(N_{\text{mask}};\; s),
\end{equation}
where $\mathcal{I}_{\text{mask}}$ denotes the set of sampled indices. Finally, the binary mask 
$\mathcal{M} \in \{0,1\}^{H \times W}$ is constructed such that
\begin{equation}
\mathcal{M}(i) =
\begin{cases}
1, & i \in \mathcal{I}_{\text{mask}} \quad \text{(masked)}, \\
0, & \text{otherwise\quad (visible)}.
\end{cases}
\end{equation}

In this way, under the premise of pixel-level masks, the final mask is generated adaptively, ensuring semantic-awareness from content-based computation and diversity from the stochasticity of sampling.

\paragraph{Optimization Objectives} Following the Bert~\cite{bert} and MAE~\cite{mae}, we choose L1 loss to supervise the masked part. The training objective of the restoration network can be written as:
\begin{equation}
\mathop{\arg\min}\limits_{\theta_r} 
\; \mathbb{E}\!\left[ 
\left\| \tilde{\mathcal{M}}\!\left( I - f(\operatorname{sg}(\mathcal{M}(I_d)), \theta_r ) \right) \right\| 
\right],
\end{equation}
where $\{I, I_d\}$ represents a pair of clean image and degraded image,  $f(\cdot, \theta_r)$ denotes a restoration network with parameters $\theta_r$, $\mathcal{M}(\cdot)$ is the mask generated by the AdaSAM strategy and $\tilde{\mathcal{M}}(\cdot)=1-\mathcal{M}(\cdot)$, $\operatorname{sg}(\cdot)$ denotes the stop-gradient operation (\ie, detach).

The training goal of the AdaSAM network is to minimize a weighted reconstruction loss defined as:
\begin{equation}
\mathop{\arg\min}\limits_{\theta_{m}} \; \mathbb{E} \left[
- \left\| \tilde{\mathcal{M}} \cdot \left( I - \operatorname{sg}\!\left(f(\mathcal{M}(I_d), \theta_r)\right) \right) \right\| \cdot \log(s)
\right],
\end{equation}
where \(s\) represents the predicted pixel-level mask score map used to weight the reconstruction error, and the logarithm operation on $s$ is applied to promote stable and smooth training. 

By optimizing the above objective functions, the AdaSAM network tends to produce masks over texture-rich, structurally complex regions that are inherently difficult to reconstruct, creating challenging objectives that encourage the restoration network to learn fine-grained reconstruction. The two networks exhibit an adversarial-like relationship in their optimization, and their joint optimization enhances both reconstruction quality and mask prediction accuracy.
% Two objective functions optimize the restoration and  AdaSAM networks adversarially to improve restoration quality with semantic-aware masks and enhance mask prediction accuracy.

\subsection{Finetuning with Mask Attribute Conductance Analysis}
\label{sec:finetuning_stage}
\begin{figure}[t]
  \centering
   \includegraphics[width=0.95\linewidth]{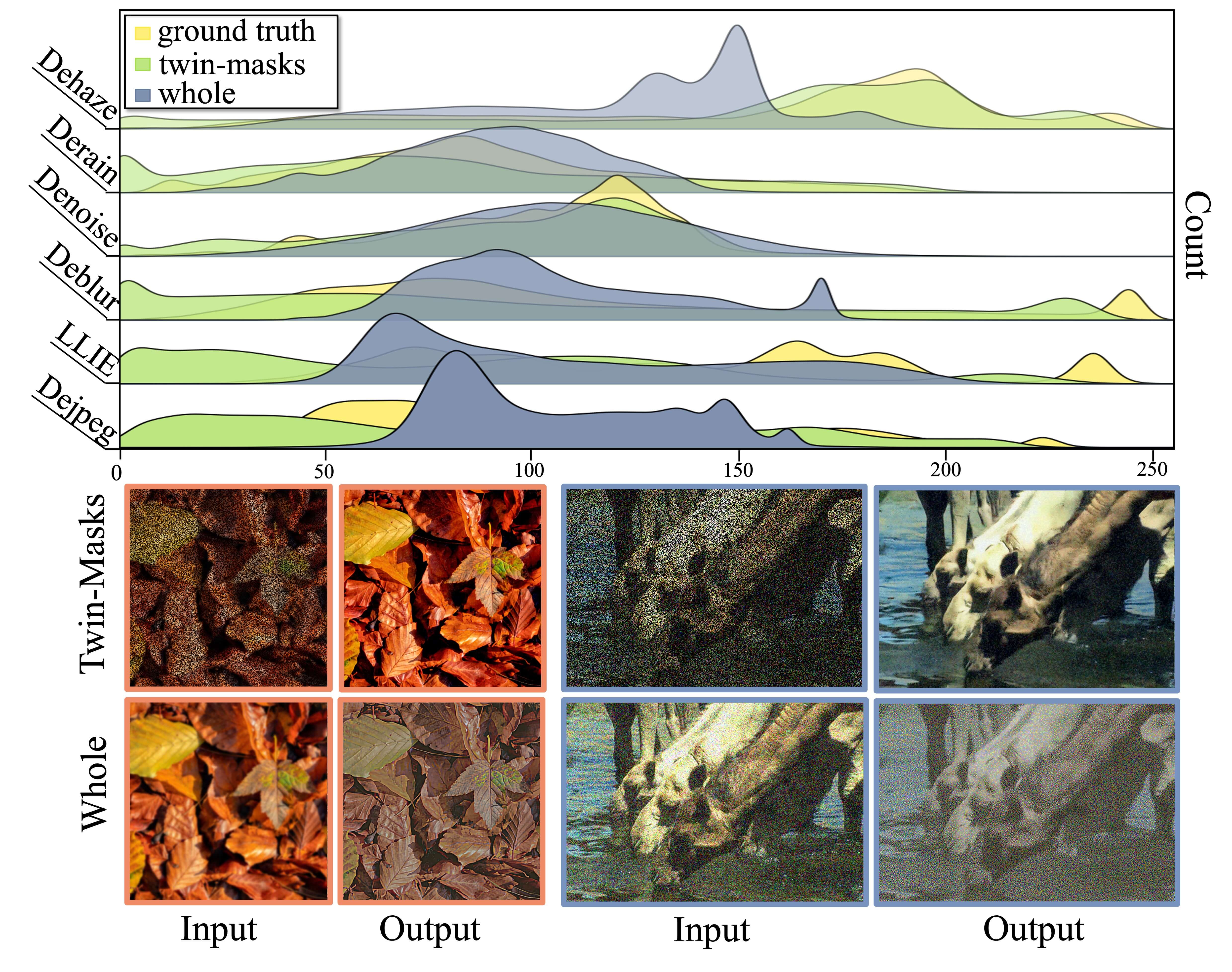}
   \caption{Effect of MIM reconstruction with different input integrity on kernel deblurring (orange border) and denoising (blue border). We also visualize the color distributions of reconstructions in various tasks above. It shows that the distribution of the reconstruction results obtained using the twin-masks method as input is closer to the real images (ground truth) compared to the results obtained using the whole input.}
   \label{fig:gap observation}
\end{figure}
\paragraph{Observation} During pre-training, the network learns rich content priors. Besides, since our AdaSAM pre-training has a similar target to subsequent low-level tasks, we do not need to change the decoder like MAE~\cite{mae} does, but just fine-tune it.
However, the incompleteness of the masked input prevents the direct use of the pre-trained model for inference, as it would result in a distribution shift in the outputs. 
As shown in Fig.~\ref{fig:gap observation}, we start by feeding the entire image into a pre-trained model with a global residual connection, leading to a color-distorted result. 
Next, we use a pair of complementary masks, referred to as twin-masks, to individually mask the image. 
Subsequently, we input both of these complementarily masked images into the network. 
By combining the pixel values predicted by each image, we generate a higher-quality image. 
This observation indicates that the hindrance to using a mask pre-trained model directly for inference lies in input incompleteness rather than the model's inability to learn the restoration function. Meanwhile, this reveals the inherent discrepancy between the generative task during the pretraining phase and the degradation fitting task during the inference phase for models with long-range connections.

Building upon this insight, we explore the possibility of minimizing the influence of disparities in data input formats via model fine-tuning. To maintain the learned priors, it is essential to retain pre-trained parameters as extensively as possible while employing the fewest but most effective layers for fine-tuning. To tackle this, we introduce the concept of mask attribution conductance, which quantifies the importance of each layer concerning the fine-tuning objective. We then identify the top-k\% most critical layers for fine-tuning. 

\paragraph{Preliminary} Before giving the definition of Mask Attribute Conductance (MAC), we briefly recall the definition of integrate gradient~\cite{ig} (IG) and neuron conductance~\cite{layerconductance} (Cond). Considering a linear path $\gamma(\alpha)=x'+\alpha(x-x')$ from base input $x'$ to target input $x$, we can attribute output change $F(x)-F(x')$ to $i$-th dimension of input/feature $x_{i}$ (\eg, a pixel) by calculating its integrate gradient, which formally as below:
\begin{equation}
  \mathrm{IG}_{i}(x):= (x_{i}-x'_{i})\cdot\int_{0}^{1} \frac{\partial F(x'+\alpha(x-x'))}{\partial x_{i}}\, d\alpha.
  \label{eq:ig}
\end{equation}
We can also attribute the output change to a specific neuron $y$ by improving IG, which involves calculating the conductance. The conductance~\cite{layerconductance} of the hidden neuron $y$ along the $\gamma (\alpha)$ is:
\begin{equation}
\begin{split}
  \mathrm{Cond}^{y}(x
  )&:= \sum_{i}(x_{i}-x'_{i})\cdot\int_{0}^{1} \frac{\partial F(x'+\alpha(x-x'))}{\partial y}\cdot \frac{\partial y}{\partial x_{i}}\, d\alpha \\ 
  & = \sum_{i}\int_{0}^{1} \frac{\partial F(\gamma(\alpha))}{\partial y}\cdot \frac{\partial y}{\partial \alpha}\, d\alpha,
  \label{eq:cond}
\end{split}
\end{equation}
Note that $(x_{i}-x'_{i})=\frac{\partial (x'+\alpha(x_{i}-x'_{i}))}{\partial \alpha}$.
Certainly, we can broaden Eq.~\eqref{eq:cond} to compute conductance when integrating along any given path $\alpha:[s,t]\rightarrow P$:
\begin{equation}
  \mathrm{GeneralCond}^{y}(x):= \sum_{i}\int_{P} \frac{\partial F(X_{i}(\alpha))}{\partial y}\cdot \frac{\partial y}{\partial \alpha}\, d\alpha,
  \label{eq:generalcond}
\end{equation}
where $X:R\rightarrow R^{n}$ is the function of the path from $x'$ to $x$, which satisfies $X(s)=x'$, $X(t)=x$. $[s,t]$ represents the domain of the path function X. 

\begin{figure}[t]{
  \centering
   \includegraphics[width=0.95\linewidth]{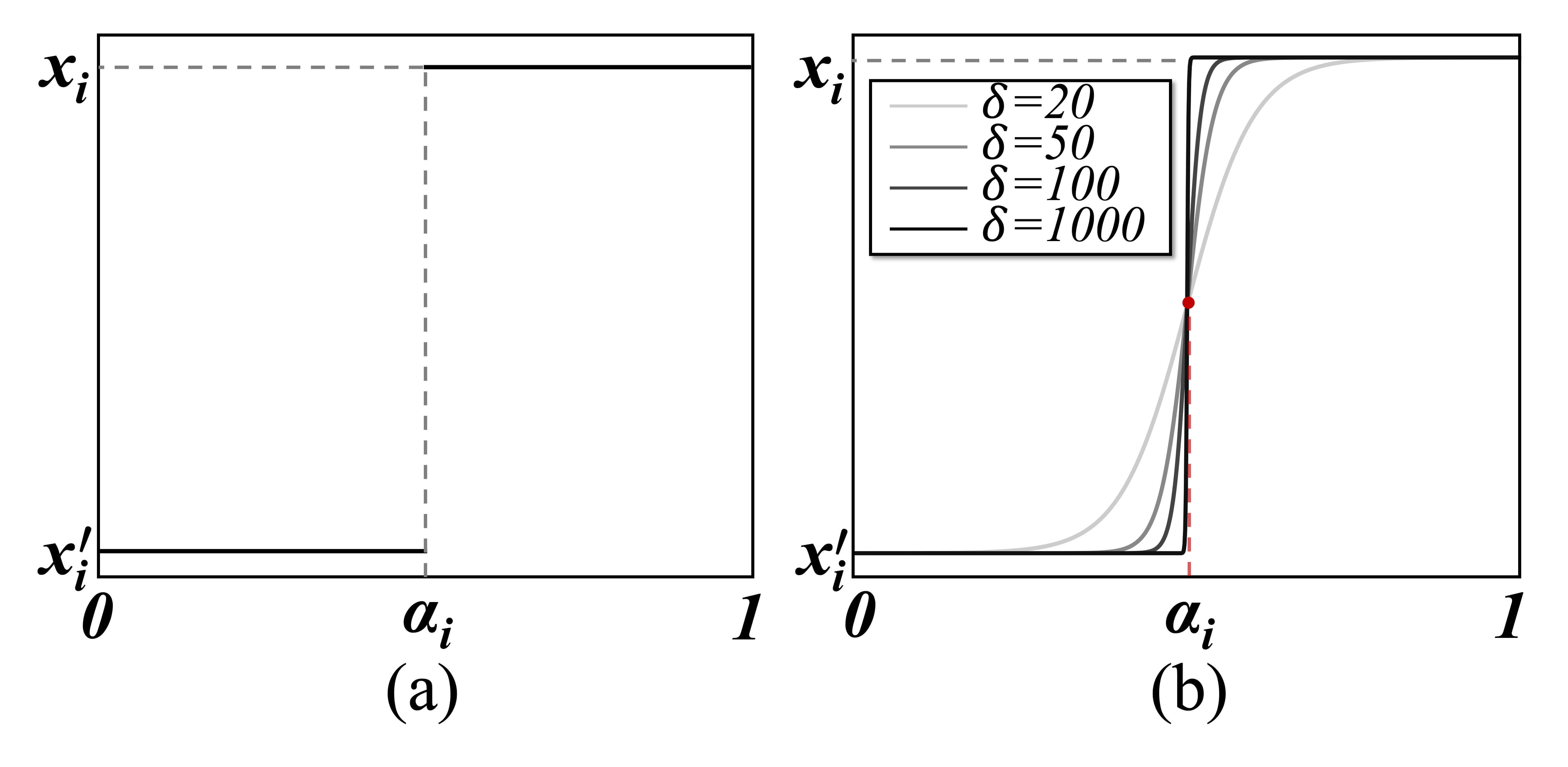}
   \caption{Illumination of (a) $X_i^{m}$ in Eq. \eqref{eq:hardmaskpath} and (b) $\tilde{X}_{i}^{m}$ in Eq. \eqref{eq:neuron softmaskpath}. }
   \label{fig:function}
}
\end{figure}

\paragraph{Finetuning with MAC} To find effective layers to finetune, we propose MAC to evaluate how effective each layer is in overcoming the gap of input integrity. Considering such a nonlinear path $\alpha: [0,1] \rightarrow P_{m}$ from zero input $x'$ to whole input $x$, which path function $X^m$ satisfies:
\begin{equation}
  X_{i}^{m}(\alpha;\alpha_i) = \begin{cases}
x'_{i}, & \alpha < \alpha_i \\
% x_i, & \alpha >= \alpha_i \\
x_i, & else \\
\end{cases},
  \label{eq:hardmaskpath}
\end{equation}
where $i$ refers to the index of pixels, $\alpha_i\in (0,1]$ is a set of parameters that indicate when each pixel gets masked. We define this path as a Mask Attribute Path (MAP). Apparently, $X^m(0)=x'$ and $X^m(1)=x$. 

However, $X^m$ is not differentiable, making it an invalid attribute path function. To solve this problem, we use a group of sigmoid-like functions  $\tilde{X}^{m}$ to approximate $X^m$:
\begin{equation}
  \tilde{X}_{i}^{m}(\alpha;\alpha_i) = \frac{(x'_{i}-x_{i})}{1+e^{-\delta(x'_{i}-\alpha_{i})}}.
  \label{eq:neuron softmaskpath}
\end{equation}

We can see that $\tilde{X}^{m}$ is very close to $X^{m}$ when $\delta$ is sufficiently large (as depicted in Fig.~\ref{fig:function}). And for each $\tilde{X}_{i}^{m}$, it will change sharply from $x'_{i}$ to $x_i$ when $\alpha$ is in the neighborhood of $\alpha_{i}$.

Here, we can define \textbf{MAC} as below:
\begin{equation}
\begin{split}
  \mathrm{MAC}^{y}(x)&:= \sum_{i}\int_{P_m} \frac{\partial F( {X}_{i}(\alpha))}{\partial y}\cdot \frac{\partial y}{\partial \alpha}\, d\alpha \\
  & \approx  \sum_{i}\int_{0}^{1} \frac{\partial F( \tilde{X}_{i}^{m}(\alpha;\alpha_i))}{\partial y}\cdot \frac{\partial y}{\partial \alpha}\, d\alpha.
  \label{eq:mac}
\end{split}
\end{equation}

In fact, a partial path is also available to attribute from a masked input $x_m$ with any mask ratio $r$ to the whole input $x$:
\begin{equation}
  \mathrm{MAC}_{r}^{y}(x)\approx  \sum_{i}\int_{1-r}^{1} \frac{\partial F( \tilde{X}_{i}^{m}(\alpha;\alpha_i))}{\partial y}\cdot \frac{\partial y}{\partial \alpha}\, d\alpha.
  \label{eq:mac_extend}
\end{equation}

In practice, we use N-steps discretization to approximate the integral form of Eq. \eqref{eq:mac_extend}, which follows 
~\cite{shrikumar2018computationally}:
\begin{equation}
\begin{split}
  \mathrm{MAC}_{r}^{y}(x) &\approx \sum_{i}\sum_{j=1}^{N} \frac{\partial F( \tilde{X}_{i}^{m}(\frac{jr}{N};\alpha_i))}{\partial y} \\
  &\cdot (F_{y}(\tilde{X}_{i}^{m}(\frac{(j+1)r}{N}))-F_{y}(\tilde{X}_{i}^{m}(\frac{jr}{N}))).
  \label{eq:mac_extend_approx}
\end{split}
\end{equation}

We compute the MAC of each layer of pre-trained networks, rank them in descending order based on their MAC values, and pick the top-$k\%$ layers for fine-tuning.
The networks are initialized with pre-trained weights, and only top-$k\%$ layers will be fine-tuned.
% More implementation details can be found in the supplementary material.

\begin{figure}[t]
  \centering
   \includegraphics[width=0.95\linewidth]{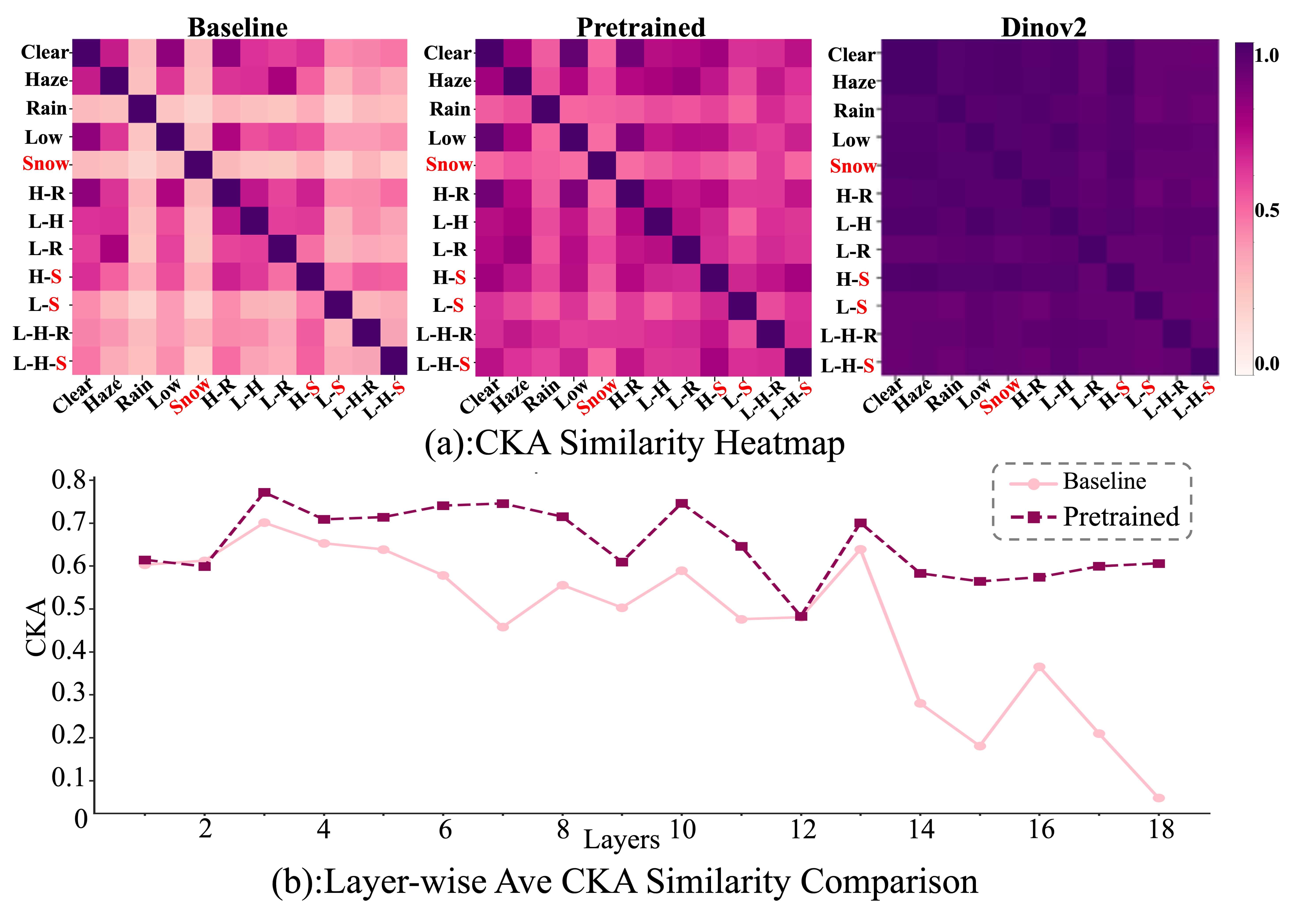}
   \caption{CKA similarity analysis across different models and their internal layer. Note: \textcolor{red}{Red} elements indicate OOD degradation type. The results demonstrate that DINOv2 possesses semantic-consistency and degradation-invariance priors, while masked pretraining models, particularly their decoders, capture degradation-invariant intrinsic image representations.}
   \label{fig:CKA}
\end{figure}

\subsection{Robust Feature Regularization}
\label{sec:dino_feature}
In this part, we first explore the robust features of DINOv2 to various degradations, as well as its necessity in compensating for the shortcomings in learning the main content during the pre-training stage. Then, we demonstrate the process of modulating features and their fusion. We emphasize that our method achieves top results not based on complex feature fusion strategies, but due to the inherent degradation robustness introduced by the pretraining scheme and the DINOv2 backbone.
\paragraph{Motivation}
% To evaluate the robustness of DINOv2 representations, we apply seven typical types of image degradation on PIES-Clean dataset~\cite{Generalization_Ability} to obtain the corresponding degraded image sets and extract deep features from the model. We compute pairwise Centered Kernel Alignment (CKA)~\cite{CKA} similarities across all eight conditions, including the clean baseline, and visualize the results as Fig.~\ref{fig:CKA}. The consistently high CKA scores indicate that DINOv2 maintains stable feature representations across various degradation types, demonstrating strong invariance in its high-level features.
According to~\cite{jin2025classic}, semantic understanding is the foundational capability for the restoration model to distinguish between the image's intrinsic texture and degradation.
To validate that the pre-trained feature extractor (\eg, DINOv2) possesses semantic understanding priors and degradation-robust priors, we extract deep features using DINOv2 on the CDD11~\cite{guo2024onerestore} dataset across 12 different degradation conditions, including the clean baseline, and compute pairwise Centered Kernel Alignment (CKA)~\cite{CKA} similarities. As shown in Fig.~\ref{fig:CKA}, DINOv2 consistently achieves high CKA scores across diverse degradation types, indicating stable and invariant high-level representations that support its role as a robust visual prior in downstream tasks. We further observe that, compared to the baseline, the masked pretraining model maintains high consistency across various degradations, including Out-Of-Distribution cases, demonstrating strong feature robustness and generalization ability.
Moreover, since the majority of target regions in the input images are occluded and thus invisible during pre-training, the model lacks sufficient exposure to the main objects.

These observations motivate the integration of DINOv2 features into the fine-tuning process, aiming to leverage their robustness to enhance image restoration performance under complex degradation scenarios and compensate for the restoration network’s limited ability to learn the main content of the image.
\begin{figure}[t]
  \centering
   \includegraphics[width=0.95\linewidth]{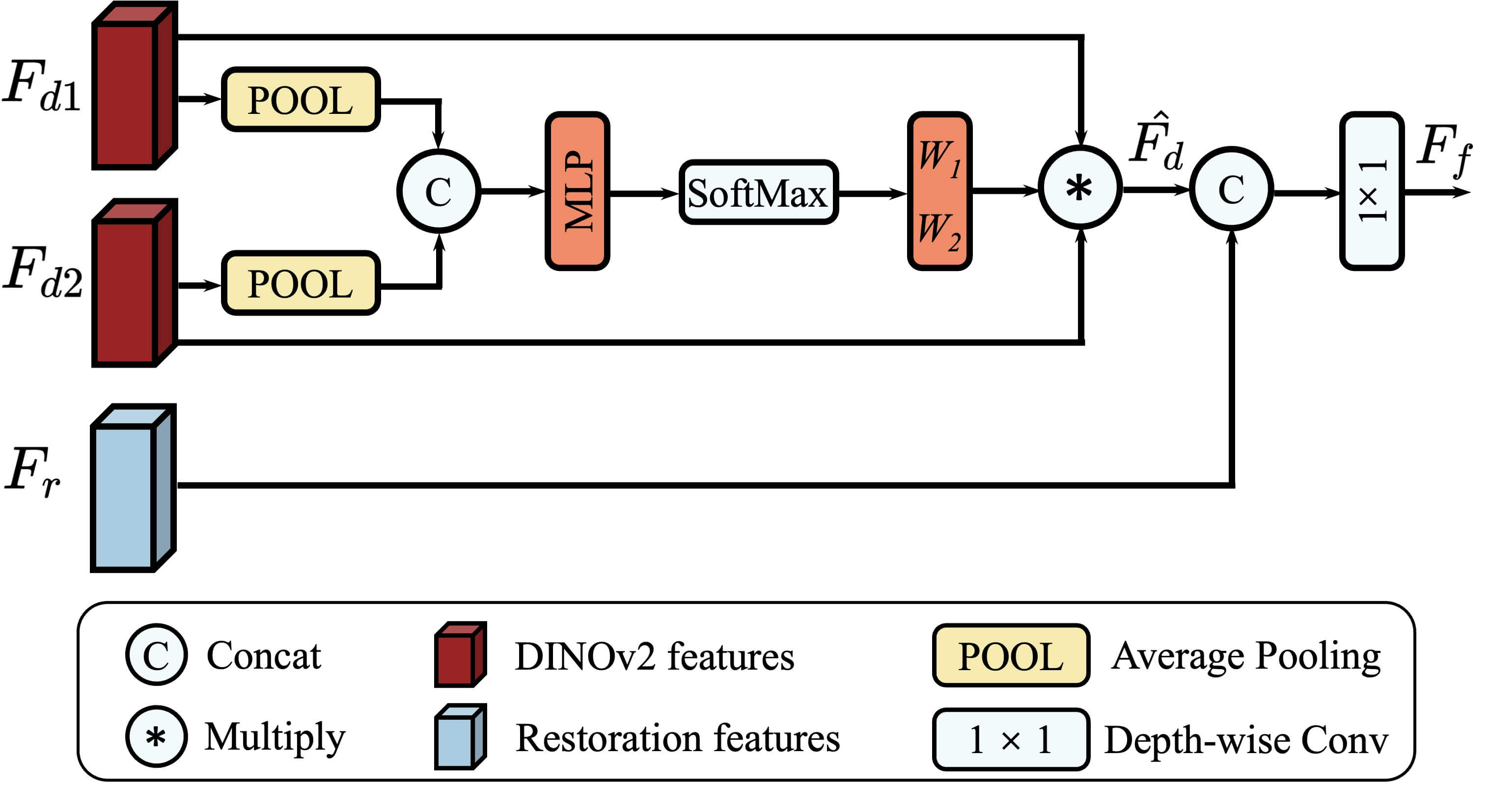}
   \caption{Overall architecture of RFR, which integrates features from Backbone and DINO-v2 branches.}
   \label{fig:feature_fusion}
\end{figure}
\paragraph{Feature Modulation and Fusion}
The overall feature modulation and fusion process is illustrated in Fig.~\ref{fig:feature_fusion}. 
The modulation module performs dynamic fusion of two hierarchical features ($F_{d_1}^i, F_{d_2}^i$) extracted from adjacent layers of DINOv2 at level $i$, where $i \in \{1, 2, 3\}$ corresponds to the shallow, middle, and deep layers, respectively. Specifically, global average pooling is applied to each feature map to extract compact semantic descriptors. These descriptors are concatenated and passed through a lightweight gating network to produce an adaptive fusion weight $w^i$. The fused DINO feature $F_d'^i$ is computed as:
\begin{align}
    w^i &= \mathrm{Gate}\left( \mathrm{Concat}\left[ \mathrm{Avg}(F_{d_1}^i), \mathrm{Avg}(F_{d_2}^i) \right] \right), \\
    \bar{F_d^i} &= w^i \cdot F_{d_1}^i + (1 - w^i) \cdot F_{d_2}^i.
\end{align}
To align the fused DINOv2 features with the restoration representation space and maintain feature balance, we design a two-layer convolutional projection module that progressively compresses the channel dimensions of DINOv2 features to ensure structural compatibility with subsequent fusion:
\begin{equation}
    \hat{F}_d^i = \mathrm{Conv}_2\left( \mathrm{ReLU}\left( \mathrm{Conv}_1(\bar{F_d^i}) \right) \right).
\end{equation}
Finally, to integrate the projected DINO features $\hat{F}_d^i$ with the corresponding restoration features $F_r^i$, we employ a straightforward yet effective fusion strategy. The two feature maps are concatenated along the channel axis and passed through a $1 \times 1$ convolutional layer to maintain the original dimensionality:
\begin{equation}
    F_f^i = F_r^i+\mathrm{Conv}_{1\times1} \left( \mathrm{Concat} \left[ \hat{F}_d^i, F_r^i \right] \right).
\end{equation}
Despite its simplicity, this fusion design proves highly effective, particularly when fine-tuning under a partially frozen configuration of the MAC module. 

\section{Experiment}
\subsection{Experiments Settings}
To demonstrate the effectiveness of the proposed \methodname{}, we conduct experiments in two different settings following previous works: \textbf{(I) 3-task image restoration}~\cite{promptir}: including dehazing, deraining, and denoising; \textbf{(II) 7-task image restoration}~\cite{qin2024restore}: additionally including motion deblurring, low-light image enhancement, kernel deblurring, and JPEG artifact removal. For each setting, we train a separate model to handle the various degradations under that setting.
\paragraph{Datasets and Metrics}  
For setting I, we adopt datasets commonly used in image restoration tasks, following established work~\cite{promptir}. For dehazing, the SOTS dataset~\cite{sots} is utilized, which provides 72,135 training samples and 500 test images. For deraining, the Rain100L~\cite{rain100} dataset is employed, which provides 200 paired clean and rainy images for training and an additional 100 pairs for testing. For denoising, the training data is composed of a combination of the BSD400~\cite{bsd400} and WED~\cite{wed} datasets, totaling 5,144 images, with the BSD68~\cite{bsd68} dataset used for evaluation. To generate noisy inputs, synthetic Gaussian noise with varying intensities ($\sigma \in \{15, 25, 50\}$) is added to the clean images.

For setting II, we combine datasets from various restoration tasks to form the training set, following~\cite{zhang2023ingredient}. For high-cost tasks where degradations are difficult to synthesize, we leverage existing paired datasets, including RESIDE~\cite{sots} for dehazing, Rain13k~\cite{fu2017clearing,li2018recurrent,li2016rain,luo2015removing,yang2020single} for deraining, GoPro~\cite{gopro} for motion deblurring, and LOL-v2~\cite{lolv2} for low-light image enhancement (LLIE). For low-cost tasks that degradations are easy to synthesize (\eg, noise, kernel blur, and JPEG artifact), we generate degraded images on the LSDIR dataset~\cite{li2023lsdir} during the training process, which involves generating Gaussian noise with random variation $\sigma\in(0,50]$, creating gaussian blurred images with a blur kernel of size $k=15$ and random $\sigma\in[0.1,3.1]$, and introducing JPEG artifacts with a random quality parameter $q\in[20,90]$.
For evaluation, we use SOTS-outdoor~\cite{sots} for dehazing, Rain13k-Test (the combination of Rain100L~\cite{rain100}, Rain100H~\cite{rain100}, Test100~\cite{test100}, Test1200~\cite{test1200}, and Test2800~\cite{test2800}) for deraining, GoPro for motion deblurring, LOL~\cite{LOL} for low-light enhancement, BSD68~\cite{bsd68} for denoising, LSDIR-val for kernel deblurring and JPEG artifact removal. Furthermore, we conducted evaluations including denoising tests with $\sigma \in \{15, 25, 50\}$, deblurring tests at $k=15$ and $\sigma=2.0$, and JPEG artifact removal tests at $q=50$.

\begin{figure}[h]
  \centering
   \includegraphics[width=0.95\linewidth]{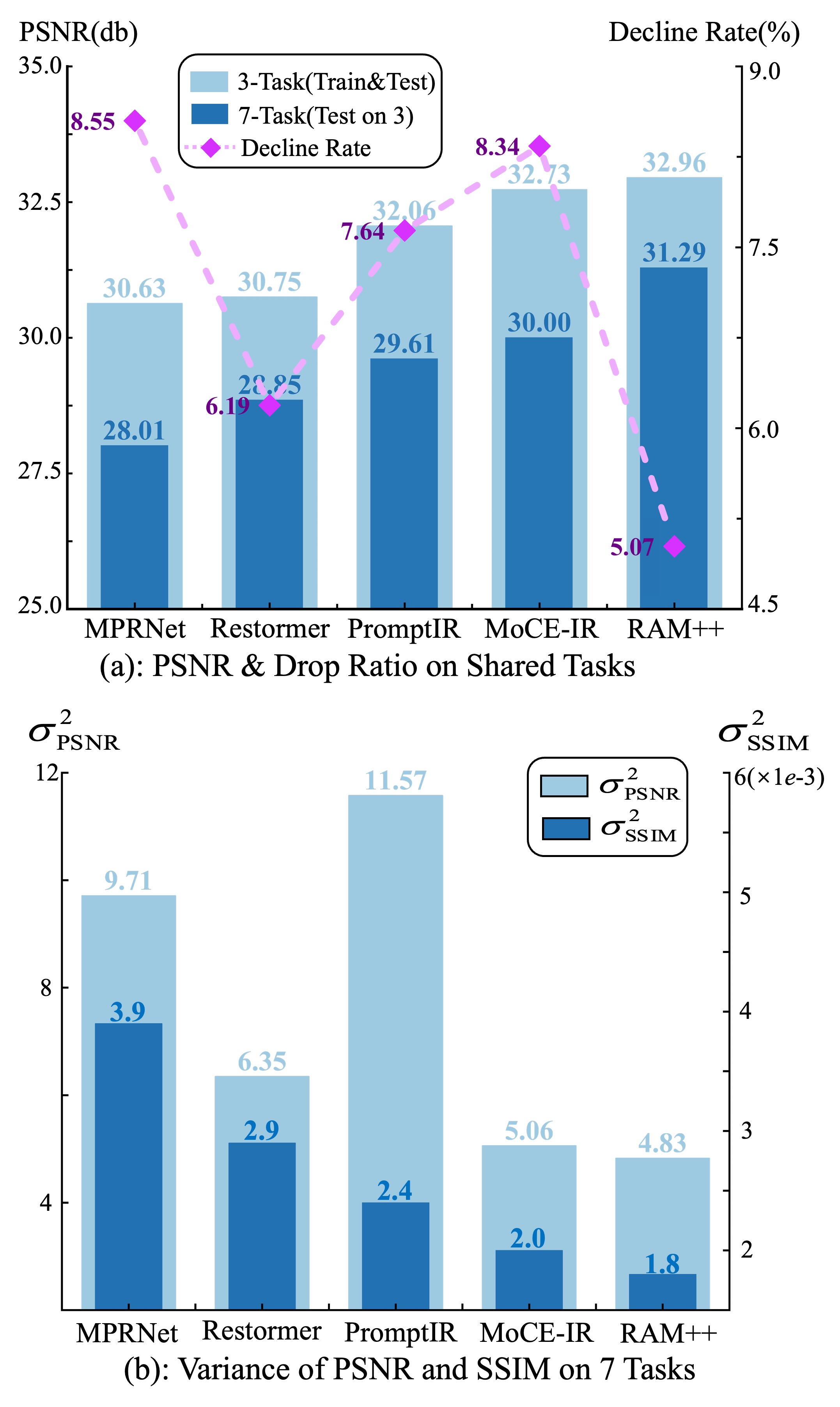}
   \caption{PSNR values and relative performance drops of different models on three shared tasks under both settings, as well as the variance of PSNR and SSIM across seven tasks. These results demonstrate that our strategy maintains performance on original tasks while achieving stronger and more balanced results across multiple tasks as the task set expands.}
   \label{fig:decline_sd}
\end{figure}

\paragraph{Implementation Details} To avoid introducing any supervision on degradation 
classification during training, we did not adopt the conference version backbone, PromptIR~\cite{promptir}, which incorporates latent degradation-aware visual prompts. Instead, we applied our proposed \methodname{} to Restormer~\cite{Zamir2021Restormer}, which is a more fundamental and general backbone framework that employs a four-level hierarchical encoder–decoder design. Notably, we eliminate the skip connections to better conform to our proposed strategy. For Setting I, the input size for \methodname{} is set to 128.   Consistent with the conference version~\cite{qin2024restore}, we set the masking ratio to 50\%. During the pre-training phase, we employ the Adam optimizer to train \methodname{} for 240 epochs, with the learning rate decaying from $2e{-4}$ to $1e{-6}$ following a cosine schedule. For AdaSAM, to stabilize training, the learning rate is decayed from $1e{-4}$ to $1e{-6}$, and the loss weight is set to $1e{-4}$. In the fine-tuning phase, we use the Adam optimizer to fine-tune the network layers obtained from the MAC analysis of \methodname{} for 30 epochs, with the learning rate decaying from $2e{-4}$ to $1e{-7}$ following a cosine schedule. 
% The batch sizes for RAM++ during the pre-training and fine-tuning phases are (4,4), respectively. 
Based on empirical observations, we select pairs of layers, specifically the 7th, 15th, 20th, 22nd, 33rd, and 39th layers, from DINOv2 to serve as representative layers for shallow-, middle-, and deep-level feature extraction. For setting II, we train \methodname{} for 300 epochs in the pre-training phase and finetune for 40 epochs.  The remaining hyperparameters are kept consistent with Setting I.

% \begin{table}[t]
% \centering
% \setlength{\tabcolsep}{1pt}
% \begin{tabular}{c|ccccc|c}
% \toprule
%  \multirow{2}{*}{Method}& \multicolumn{5}{c|}{Out-Of-Distribution} &  In-Distribution \\
% & Salt & Poisson & Pepper & Speckle & Ave & Ave \\
% \midrule
% Baseline & 26.35 & 10.92 & 24.87 & 16.96 & 19.78 & 31.07 \\
% \methodname$_{\text{100\%}}$ & 27.84 & 13.05 & 26.10 & 19.13 & 21.53 & 31.10 \\
% \bottomrule
% \end{tabular}
% \caption{Performance comparison on in-distribution and out-of-distribution noise types.}
% \label{tab:ood_ind}
% \end{table}

\begin{table*}[t]
    \centering
     \caption{Quantitative comparison on three challenging image restoration tasks, including dehazing, deraining, and denoising.  \textcolor{blue}{blue} and \textcolor{teal}{teal} indicate the best and second-best results, respectively.}
     \vspace{3pt} 
    \label{tab:allinone_3}
    \begin{tabular}{c|ccccc|c@{}}
    \hline
        \multirow{2}{*}{Method} & 
        \multirow{2}{*}{SOTS~\cite{sots}} & 
        \multirow{2}{*}{Rain100L~\cite{rain100}} & 
        \multicolumn{3}{c|}{BSD68 dataset~\cite{bsd68}} & 
        \multirow{2}{*}{Average} \\ 
        ~ & ~ & ~ & $\sigma$ = 15 & $\sigma$ = 25 & $\sigma$ = 50 & ~ \\ \hline
       Restormer~\cite{Zamir2021Restormer} 
& 27.78/0.958 
& 33.72/0.865 
& 33.72/0.865 
& 30.67/0.865 
& 27.63/0.792 
& 30.75/0.901 \\

MPRNet~\cite{mehri2021mprnet} 
& 28.00/0.958 
& 33.86/0.958 
& 33.27/0.920 
& 30.76/0.871 
& 27.29/0.761 
& 30.63/0.894 \\

NAFNet~\cite{nafnet} 
& 24.11/0.960 
& 33.64/0.956 
& 33.18/0.918 
& 30.47/0.865 
& 27.12/0.754 
& 29.67/0.844 \\

DL~\cite{DL} 
& 26.92/0.931 
& 32.62/0.931 
& 33.05/0.914 
& 30.41/0.861 
& 26.90/0.740 
& 29.98/0.875 \\

AirNet~\cite{airnet} 
& 27.94/0.962 
& 34.90/0.967 
& 33.92/\textcolor{teal}{0.933} 
& 31.26/0.888 
& 28.00/0.797 
& 31.20/0.910 \\

PromptIR~\cite{promptir} 
& 30.58/0.974 
& 36.37/0.972 
& 33.98/\textcolor{teal}{0.933} 
& 31.31/0.888 
& 28.06/0.799 
& 32.06/0.913 \\

InstructIR-3D~\cite{instructir} 
& 30.22/0.959 
& 37.98/0.978 
& \textcolor{teal}{34.15}/\textcolor{teal}{0.933} 
& \textcolor{teal}{31.52}/\textcolor{teal}{0.890} 
& \textcolor{blue}{28.30}/\textcolor{blue}{0.804} 
& 32.43/0.913 \\

Art$_{\mathrm{AirNet}}$~\cite{art} 
& 30.56/0.977 
& 37.74/0.981 
& 34.02/\textcolor{blue}{0.934} 
& 31.37/\textcolor{teal}{0.890} 
& 28.12/\textcolor{teal}{0.802} 
& 32.36/\textcolor{teal}{0.917} \\

Art$_{\mathrm{PromptIR}}$~\cite{art} 
& 30.83/\textcolor{teal}{0.979} 
& 37.94/0.982 
& 34.06/\textcolor{blue}{0.934} 
& 31.42/\textcolor{blue}{0.891} 
& 28.14/0.801 
& 32.49/\textcolor{teal}{0.917} \\

GridFormer~\cite{gridformer} 
& 30.37/0.970 
& 37.15/0.972
& 33.93/0.931 
& 31.37/0.887 
& 28.11/0.801 
& 32.19/0.912 \\

RAM$_{\mathrm{Restormer}}$~\cite{qin2024restore}
& 30.26/0.975
& 37.89/0.982
& \textcolor{teal}{34.15}/0.932
& 31.51/0.888
& 28.24/0.800
& 32.41/0.915 \\

MoCE-IR~\cite{moceir}  
& \textcolor{teal}{31.12}/\textcolor{teal}{0.979} 
& \textcolor{teal}{38.72}/\textcolor{teal}{0.984} 
& 34.13/0.932 
& 31.47/0.888 
& 28.20/0.800 
& \textcolor{teal}{32.73}/\textcolor{teal}{0.917} \\ \hline

\methodname$_{\text{30\%}}$ 
& 31.10/0.978 
& 37.78/0.981 
& 34.09/0.932 
& 31.49/0.889 
& 28.25/0.801 
& 32.54/0.916 \\

\methodname$_{\text{100\%}}$ 
& \textcolor{blue}{31.90}/\textcolor{blue}{0.980} 
& \textcolor{blue}{38.87}/\textcolor{blue}{0.985} 
& \textcolor{blue}{34.18}/\textcolor{teal}{0.933} 
& \textcolor{blue}{31.55}/0.889 
& \textcolor{teal}{28.29}/\textcolor{teal}{0.802} 
& \textcolor{blue}{32.96}/\textcolor{blue}{0.918} \\ \hline
    \end{tabular}
\end{table*}

\begin{table*}
  \centering
  \renewcommand\arraystretch{1.1}
  \caption{Quantitative comparison on seven challenging image restoration tasks, including dehazing, deraining, denoising, motion deblurring, low-light image enhancement (LLIE), kernel deblurring, and JPEG artifact removal.  \textcolor{blue}{blue} and \textcolor{teal}{teal} indicate the best and second-best results, respectively.}
  \vspace{3pt} 
  \label{tab:allinone}
  \resizebox{\textwidth}{!}{
  \begin{tabular}{c|ccccccc|c@{}}
    \toprule
     \multirow{2}{*}{Method}
     & SOTS~\cite{sots} & Rain13k-Test~\cite{degae} & BSD68~\cite{bsd68} & GoPro~\cite{gopro} & LOL~\cite{LOL} & LSDIR-Blur~\cite{li2023lsdir} & LSDIR-JPEG~\cite{li2023lsdir} & Average \\
      & 
  PSNR$\uparrow$/SSIM$\uparrow$& PSNR$\uparrow$/SSIM$\uparrow$& PSNR$\uparrow$/SSIM$\uparrow$ & PSNR$\uparrow$/SSIM$\uparrow$ & PSNR$\uparrow$/SSIM$\uparrow$ & PSNR$\uparrow$/SSIM$\uparrow$ & PSNR$\uparrow$/SSIM$\uparrow$ & PSNR$\uparrow$/SSIM$\uparrow$  \\
    \midrule
Restormer~\cite{Zamir2021Restormer} & 22.89/0.9172 & 27.05/0.8469 & \textcolor{blue}{30.95}/\textcolor{blue}{0.8657} & 27.46/0.8497 & 23.65/0.8458 & 19.60/0.3658 & \textcolor{blue}{30.46}/\textcolor{blue}{0.9141} & 26.01/0.8007 \\
MPRNet~\cite{mehri2021mprnet} & 25.23/0.9463 & 25.36/0.8068 & 29.83/0.8317 & 25.90/0.7949 & 22.29/0.8170 & 25.68/0.8281 & 28.96/0.8865 & 26.18/0.8445 \\
NAFNet~\cite{nafnet} & 25.74/0.9445 & 24.65/0.7877 & 30.37/0.8540 & 25.53/0.7909 & 21.50/0.8104 & 29.08/0.9130 & 29.09/0.8955 & 26.57/0.8566 \\
DL~\cite{DL} & 21.16/0.9042 & 19.56/0.6508 & 16.15/0.5861 & 17.63/0.5862 & 19.26/0.7777 & 17.98/0.6121 & 19.55/0.6965 & 18.75/0.6877 \\
TAPE~\cite{liu2022tape} & 25.14/0.9319 & 23.66/0.7818 & 30.11/0.8354 & 25.97/0.7962 & 18.95/0.7632 & 24.26/0.7654 & 29.28/0.8965 & 25.34/0.8243 \\
AirNet~\cite{airnet} & 21.66/0.8366 & 20.21/0.6402 & 27.99/0.7250 & 23.36/0.7503 & 16.65/0.6708 & 23.84/0.7358 & 24.36/0.8020 & 22.58/0.7372 \\
SwinIR~\cite{liang2021swinir} & 27.29/0.9622 & 25.32/0.8258 & 30.65/0.8540 & 26.61/0.8125 & 18.66/0.8048 & 27.82/0.8839 & 30.13/0.9071 & 26.64/0.8643 \\
PromptIR~\cite{promptir} & 28.70/0.9659 & 27.46/0.8585 & 30.84/0.8625 & 27.71/0.8565 & 21.19/0.8356 & \textcolor{teal}{31.01}/\textcolor{teal}{0.9385} & 30.30/0.9117 & 28.17/0.8899 \\
MoCE-IR~\cite{moceir} & 30.16/0.9719 & 27.86/0.8608 & 30.57/0.8556 & 27.70/0.8506 & 24.00/0.8446 & 27.96/0.8837 & 29.88/0.9049 & 28.30/0.8817 \\
HOGFormer~\cite{hogformer} & 29.21/0.9685 & 27.09/0.8415 & 30.61/0.8551 & 27.64/0.8477 & 21.38/0.8273 & \textcolor{blue}{32.02}/\textcolor{blue}{0.9480} & 30.20/0.9092 & 28.31/0.8853 \\
RAM$_{\mathrm{SwinIR}}$~\cite{qin2024restore} & 28.47/0.9689 & 26.31/0.8486 & 30.83/0.8611 & 26.89/0.8200 & 21.62/0.8291 & 26.66/0.8514 & 30.22/0.9096 & 27.28/0.8698 \\
RAM$_{\mathrm{PromptIR}}$~\cite{qin2024restore} & 29.64/0.9695 & 28.47/0.8751 & 30.86/0.8624 & \textcolor{teal}{28.02}/\textcolor{teal}{0.8592} & 24.46/0.8581 & 29.57/0.9179 & 30.33/0.9119 & 28.76/\textcolor{teal}{0.8935} \\
RAM$_{\mathrm{Restormer}}$~\cite{qin2024restore} & 29.97/0.9721 & 27.97/0.8666 & \textcolor{teal}{30.88}/0.8631 & 27.83/0.8582 & 24.15/\textcolor{teal}{0.8613} & 29.50/0.9153 & 30.37/0.9127 & 28.67/0.8928 \\
\midrule
\methodname$_{\text{30\%}}$ & \textcolor{teal}{31.67}/\textcolor{teal}{0.9760} & \textcolor{teal}{28.52}/\textcolor{teal}{0.8782} & 30.82/\textcolor{teal}{0.8654} & 27.61/0.8489 & \textcolor{teal}{24.86}/0.8545 & 28.45/0.8926 & 30.19/0.9107 & \textcolor{teal}{28.88}/0.8895 \\
\methodname$_{\text{100\%}}$ & \textcolor{blue}{31.91}/\textcolor{blue}{0.9782} & \textcolor{blue}{29.53}/\textcolor{blue}{0.8919} & \textcolor{blue}{30.95}/0.8652 & \textcolor{blue}{28.19}/\textcolor{blue}{0.8621} & \textcolor{blue}{25.23}/\textcolor{blue}{0.8619} & 29.96/0.9219 & \textcolor{teal}{30.43}/\textcolor{teal}{0.9139} & \textcolor{blue}{29.46}/\textcolor{blue}{0.8993} \\
  \midrule
  \end{tabular}}
\end{table*}

\subsection{Comparisons}
\paragraph{General Evaluation}
We compare our method with four general image restoration approaches~\cite{nafnet,mehri2021mprnet,Zamir2021Restormer,liang2021swinir} and ten all-in-one restoration methods~\cite{DL,liu2022tape,airnet,promptir,moceir,hogformer,instructir,art,gridformer,qin2024restore}.
To ensure fairness, all methods use the same number of supervised pixels as in our pre-training stage.

As shown in Tab.~\ref{tab:allinone_3}, our method achieves the best or comparable results across all tasks. The model fine-tuned with $30\%$ of the data achieves strong performance while retaining pre-trained knowledge. In comparison, the model fine-tuned with $100\%$ of the data shows a $0.23$dB improvement over the current best-performing model. Tab.~\ref{tab:allinone} presents the results of seven tasks in the specified setting. With $30\%$ fine-tuning, our method outperforms the second-best method by $0.12$dB. When fully fine-tuned ($100\%$), the improvement reaches $0.70$dB across seven different tasks.
Notably, \methodname{} brings significant improvements in dehazing, deraining, and low-light enhancement.

Meanwhile, Fig.~\ref{fig:decline_sd} illustrates that as the number of tasks increases, the \methodname{} strategy not only enhances overall performance but also minimizes decline in the original tasks, with the performance decline limited to just $5.07$\%. And our model inevitably maintains balanced performance across them, exhibiting the lowest variance of $4.83$ across the seven tasks. This is driven by the synergy between the strong generative prior from masked image modeling and the semantic-consistency and degradation-invariance priors of DINOv2.

% Tab.~\ref{tab:denoise_test} presents quantitative results for denoising under different noise levels.
% Both \methodname-SwinIR and \methodname-PromptIR outperform their original versions across all settings.
\begin{figure*}[t]
  \centering
   \includegraphics[width=0.88\linewidth]
   {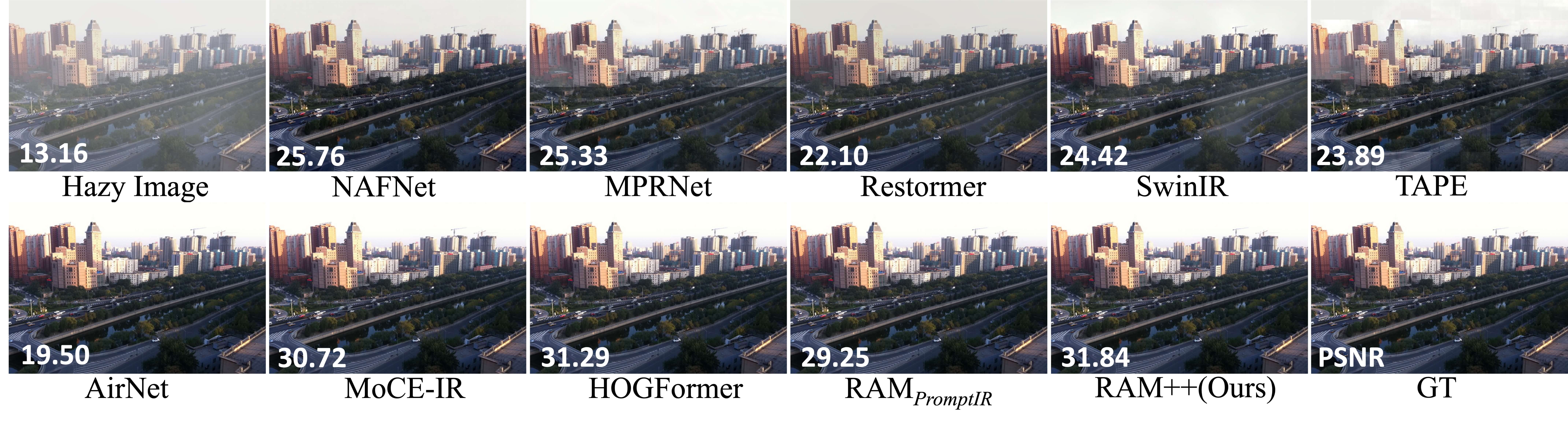}
   \caption{Dehaze visual comparison on SOTS dataset. Zoom in for details.}
   \label{fig:haze_result}
\end{figure*}

\begin{figure*}[t]
  \centering
   \includegraphics[width=0.88\linewidth]
   {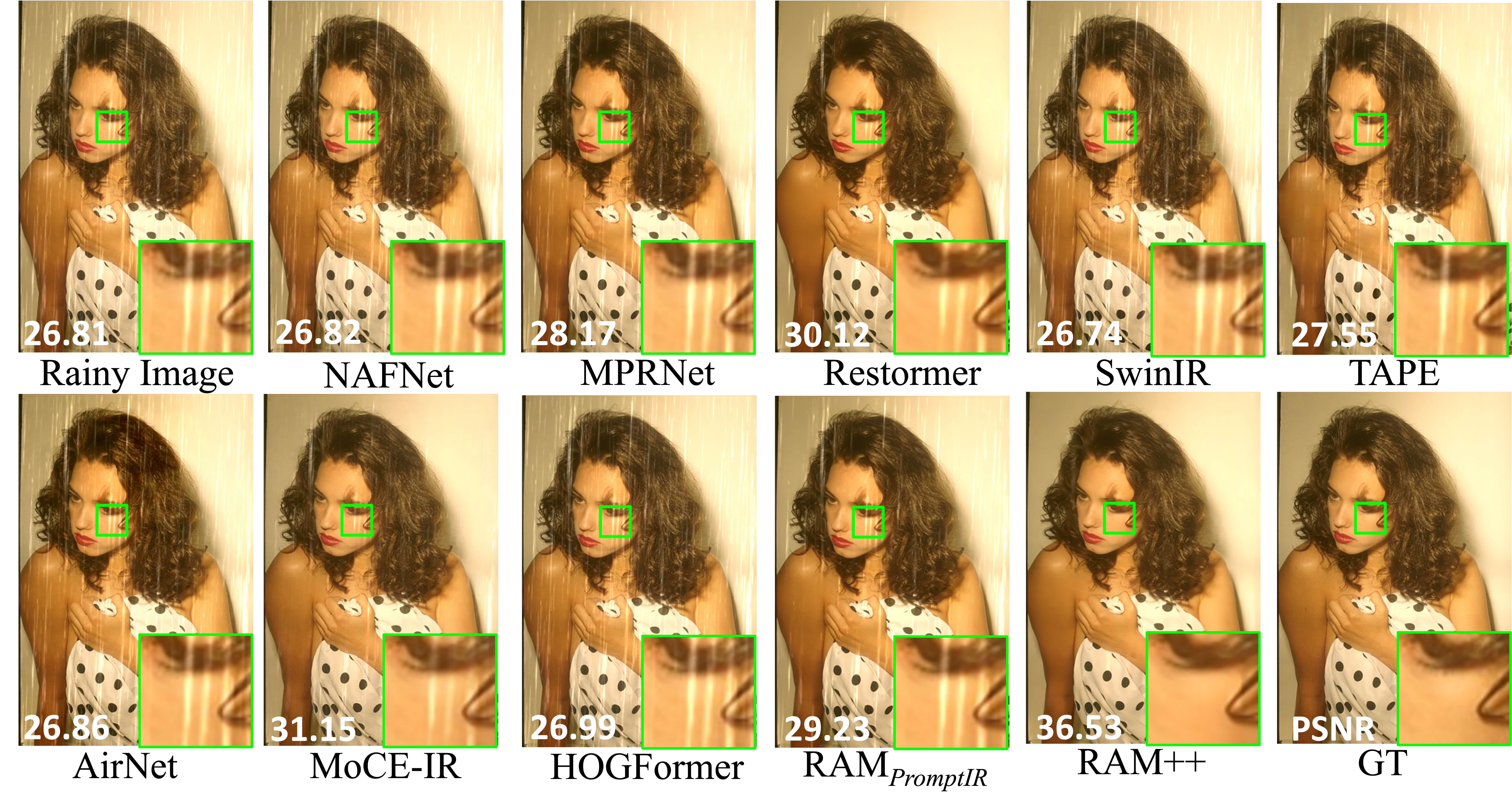}
   \caption{Derain visual comparsion on Rain13k-Test dataset. Zoom in for details.}
   \label{fig:rain_result}
\end{figure*}

\begin{figure*}[h]
  \centering
   \includegraphics[width=0.88\linewidth]{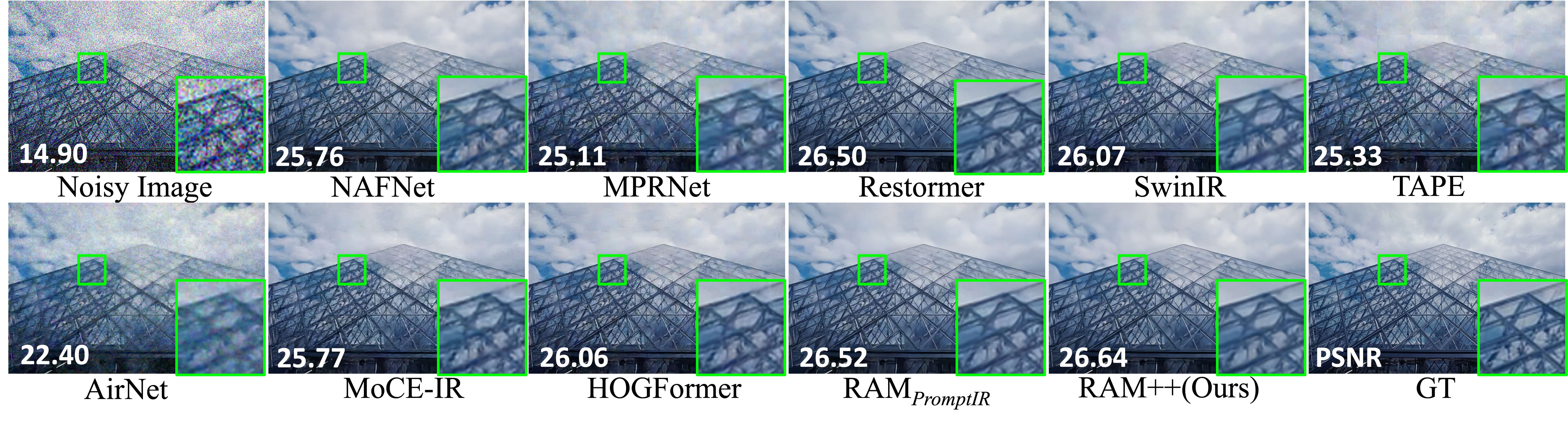}
      \setlength{\abovecaptionskip}{2pt}
   \caption{Denoising visual comparison on CBSD68 dataset. Zoom in for details.}
   \label{fig:noise_result}
\end{figure*}

\begin{figure*}[h]
  \centering
   \includegraphics[width=0.88\linewidth]{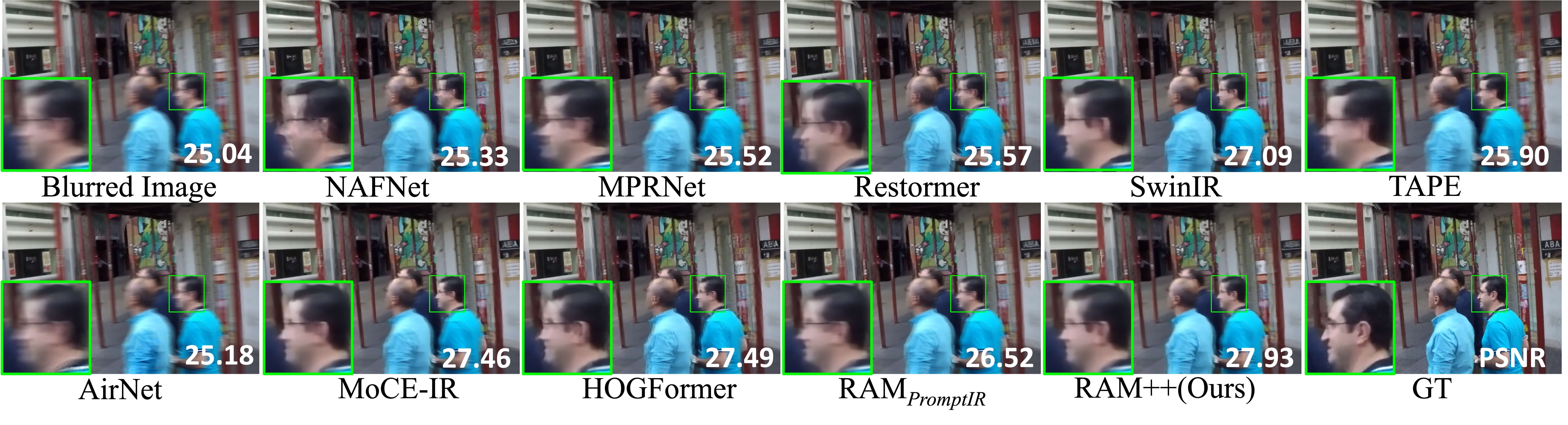}
      \setlength{\abovecaptionskip}{2pt}
   \caption{Motion deblur visual comparison on GoPro dataset. Zoom in for details.}
   \label{fig:blur_result}
\end{figure*}

\begin{figure*}[h]
  \centering
   \includegraphics[width=0.88\linewidth]{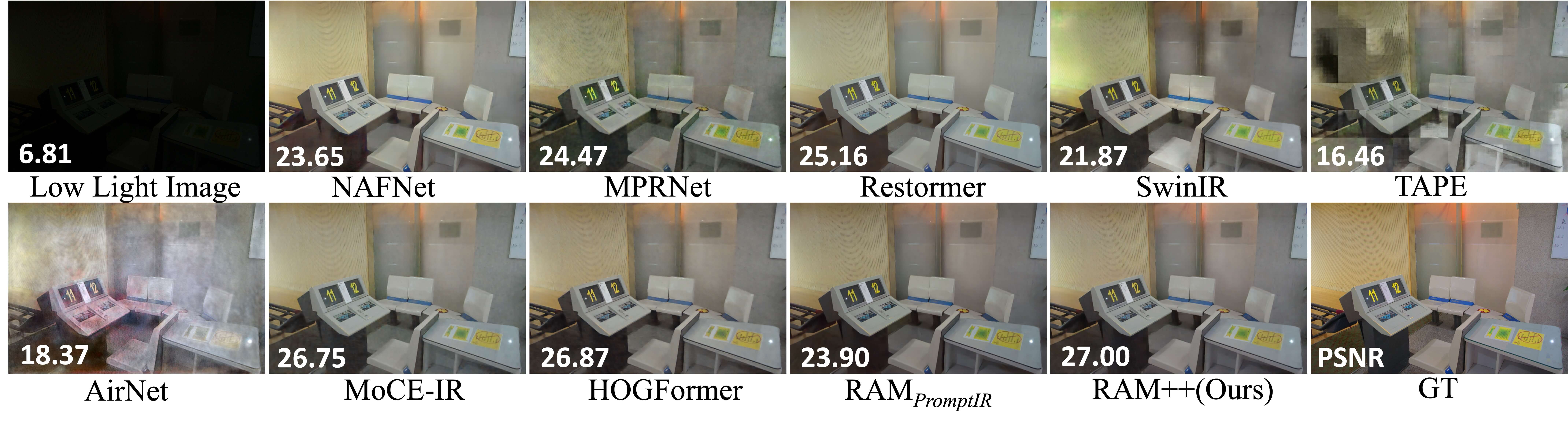}
   \caption{LLIE visual comparison on LOL dataset. Zoom in for details.}
   \label{fig:lowlight_result}
\end{figure*}

\cref{fig:haze_result,fig:rain_result,fig:noise_result,fig:blur_result,fig:lowlight_result} show qualitative comparisons on various datasets under setting II.
In Fig.~\ref{fig:haze_result}, our method delivers better dehazing (right region) and more accurate exposure correction (sky).
In the deraining task (Fig.~\ref{fig:rain_result}), our model removes rain streaks more effectively and recovers clearer textures in occluded areas. 
For denoising (Fig.~\ref{fig:noise_result}) and deblurring (Fig.~\ref{fig:blur_result}), we obtain sharper images with fewer artifacts.
In the low-light enhancement task (Fig.~\ref{fig:lowlight_result}), our method provides better color correction (\eg, the pale yellow wall behind) and improved exposure handling.
% The qualitative effects of kernel deblurring (Fig.~\ref{fig:kernel-blur_result}) and JPEG artifact removal (Fig.~\ref{fig:jpeg_result}) also illustrate the superiority of our method.

\begin{table*}[t]
\renewcommand\arraystretch{1.1}
\centering
\caption{Quantitative comparison of PSNR for models trained under Setting 2 (7-task) on the CDD-11 dataset~\cite{guo2024onerestore}, using four single-task, five dual-task, and two triple-task configurations. Note: \textcolor{red}{Red} entries indicate OOD (out-of-distribution) degradation types.}
\vspace{3pt} 
\label{tab:cdd11_test}
\resizebox{\textwidth}{!}{ % 缩放到版心宽度
\begin{tabular}{c|cccc|ccccc|cc|c} 
    \toprule
    \multirow{2}{*}{Method}  & \multicolumn{4}{c}{CDD11-Single}  & \multicolumn{5}{c}{CDD11-Double}& \multicolumn{2}{c}{CDD11-Triple}&\multirow{2}{*}{Average}\\
    &Low(L)&Haze(H)&Rain(R)&\textcolor{red}{Snow(S)}&L+H&L+R&L+\textcolor{red}{S}&H+R&H+\textcolor{red}{S}&L+H+R&L+H+\textcolor{red}{S}&\\
    \midrule
Airnet~\cite{airnet} & 12.04 & 16.84 & 22.35 & 22.64 & 14.93 & 14.77 & 13.56 & 15.41 & 15.44 & 15.30 & 14.65 & 16.17 \\
Restormer~\cite{Zamir2021Restormer} & 12.15 & 16.17 & 25.06 & 23.66 & 15.01 & 14.92 & 13.68 & 15.33 & 14.93 & 15.33 & 14.67 & 16.45 \\
SwinIR~\cite{liang2021swinir} & \textcolor{teal}{12.24} & 16.97 & 22.19 & 23.44 & \textcolor{teal}{15.05} & \textcolor{blue}{15.00} & \textcolor{blue}{13.76} & 16.25 & 15.69 & 15.36 & \textcolor{teal}{14.70} & 16.42 \\
PromptIR~\cite{promptir} & 12.14 & 16.53 & 26.27 & 23.65 & 15.00 & 14.64 & 13.68 & 16.09 & 15.17 & 15.38 & 14.68 & 16.66 \\
HOGFormer~\cite{hogformer} & 12.12 & 15.71 & 22.92 & 22.18 & 14.94 & 14.91 & 13.69 & 16.87 & 14.69 & 15.36 & 14.65 & 16.19 \\
MoCE-IR~\cite{moceir} & 12.14 & 15.70 & 24.92 & 20.22 & 14.99 & \textcolor{teal}{14.94} & \textcolor{blue}{13.76} & 15.32 & 13.60 & 15.36 & 14.58 & 15.96 \\
RAM$_{\mathrm{PromptIR}}$~\cite{qin2024restore} & 12.13 & 17.03 & \textcolor{blue}{28.21} & \textcolor{blue}{23.76} & 14.97 & 14.46 & 13.68 & 16.11 & 15.49 & 15.35 & 14.65 & 16.90 \\
RAM$_{\mathrm{Restormer}}$~\cite{qin2024restore} & 12.15 & \textcolor{teal}{18.11} & 26.82 & \textcolor{teal}{23.73} & 14.98 & 14.67 & 13.69 & \textcolor{teal}{17.11} & \textcolor{teal}{16.05} & \textcolor{blue}{15.41} & 14.65 & \textcolor{teal}{17.03} \\
\methodname$_{\text{100\%}}$ & \textcolor{blue}{12.26} & \textcolor{blue}{19.07} & \textcolor{teal}{27.37} & 22.78 & \textcolor{blue}{15.11} & 14.88 & \textcolor{teal}{13.71} & \textcolor{blue}{17.19} & \textcolor{blue}{16.68} & \textcolor{teal}{15.40} & \textcolor{blue}{14.71} & \textcolor{blue}{17.20} \\
    \bottomrule
\end{tabular}
 }
\end{table*}

\paragraph{Extreme Scenarios and Mixed Degradation Evaluation}
To simulate more realistic degradation scenarios, following recent studies~\cite{guo2024onerestore}, multiple types of degradation are combined within a single image to create more challenging test cases. Unlike those works that retrain the new models, we evaluate various models from Setting II (7-task) directly. Given the significant domain gap between the CDD11 test set and the datasets used for training, we treat previously seen single-task degradations as restoration tasks under extreme conditions, alongside seven types of mixed degradation tasks. As shown in Tab.~\ref{tab:cdd11_test}, our method achieves a $0.17$dB improvement over previous all-in-one image restoration approaches, while also demonstrating strong domain generalization and robust performance under mixed degradations.

\begin{table*}[t]
\renewcommand\arraystretch{1.1}
\centering
\caption{Quantitative denoising results at different noises on BSD68 and Urban100 datasets in terms of PSNR. }
\vspace{3pt} 
\label{tab:denoise_test}
\resizebox{\textwidth}{!}{ % 缩放到版心宽度
\begin{tabular}{c|cccc|cccc|cccc} % 在ID和OOD之间加了竖线
    \toprule
     \multirow{3}{*}{Method}  & \multicolumn{8}{c|}{In-Distribution}  & \multicolumn{4}{c}{Out-Of-Distribution}\\
     & \multicolumn{4}{c|}{BSD68~\cite{bsd68}}  & \multicolumn{4}{c|}{Urban100~\cite{urban100}} & \multicolumn{4}{c}{Urban100~\cite{urban100}} \\
      & $\sigma=15$ & $\sigma=25$ & $\sigma=50$ & Avg & $\sigma=15$ & $\sigma=25$ & $\sigma=50$ & Avg & Poisson & Pepper & Speckle & Avg \\
    \midrule
NAFNet~\cite{nafnet}
& 33.22 & 30.59 & 27.30 & 30.37 & 32.67 & 30.21 & 26.97 & 29.92 & \textcolor{teal}{13.00} & 25.70 & 18.33 & \textcolor{teal}{19.01} \\

MPRNet~\cite{mehri2021mprnet}
& 32.73 & 30.11 & 26.65 & 29.83 & 32.06 & 29.46 & 25.77 & 29.10 & 9.90 & \textcolor{blue}{26.46} & 17.21 & 17.86 \\

Restormer~\cite{Zamir2021Restormer}
& \textcolor{blue}{33.79} & \textcolor{teal}{31.17} & \textcolor{blue}{27.90} & \textcolor{blue}{30.95} & \textcolor{blue}{33.83} & \textcolor{teal}{31.40} & \textcolor{teal}{27.99} & \textcolor{teal}{31.07} & 10.92 & 24.87 & 16.96 & 17.59 \\

TAPE~\cite{liu2022tape}
& 33.10 & 30.37 & 26.86 & 30.11 & 32.59 & 29.93 & 26.19 & 29.57 & 9.99 & 23.24 & 16.61 & 16.61 \\

AirNet~\cite{airnet}
& 31.63 & 28.83 & 23.52 & 27.99 & 29.79 & 26.90 & 21.35 & 26.01 & 4.80 & 20.78 & 5.32 & 10.30 \\

SwinIR~\cite{liang2021swinir}
& 33.53 & 30.89 & 27.54 & 30.65 & 33.50 & 30.99 & 27.37 & 30.62 & 12.81 & 10.60 & \textcolor{blue}{20.09} & 14.50 \\

MoCE-IR~\cite{moceir}
& 33.43 & 30.80 & 27.49 & 30.57 & 33.06 & 30.52 & 26.95 & 30.18 & 11.48 & 24.87 & 16.86 & 17.74 \\

HOGFormer~\cite{hogformer}
& 33.53 & 30.85 & 27.45 & 30.61 & 33.44 & 30.90 & 27.33 & 30.56 & 11.09 & 23.98 & 16.10 & 17.06 \\

RAM$_{\mathrm{PromptIR}}$~\cite{qin2024restore}
& 33.70 & 31.08 & 27.79 & 30.86 & 33.70 & 31.30 & 27.92 & 30.97 & 11.26 & 25.23 & 16.94 & 17.81 \\

RAM$_{\mathrm{Restormer}}$~\cite{qin2024restore}
& 33.73 & 31.11 & \textcolor{teal}{27.81} & \textcolor{teal}{30.88} & 33.76 & 31.36 & 27.98 & 31.03 & 9.80 & 22.31 & 15.13 & 15.75 \\

\methodname$_{\text{100\%}}$
& \textcolor{teal}{33.77} & \textcolor{blue}{31.18} & \textcolor{blue}{27.90} & \textcolor{blue}{30.95} & \textcolor{teal}{33.79} & \textcolor{blue}{31.41} & \textcolor{blue}{28.09} & \textcolor{blue}{31.10} & \textcolor{blue}{13.05} & \textcolor{teal}{26.10} & \textcolor{teal}{19.13} & \textcolor{blue}{19.43} \\

    \bottomrule
\end{tabular}
}
\end{table*}

\paragraph{Out-of-Distribution Degradation Evaluation}
To evaluate the model's generalization ability under Out-of-Distribution (OOD) degradations, we test it on three different noise types in the Urban~\cite{urban100} dataset and underwater image enhancement capability in the UIEB~\cite{UIEB} dataset.

\begin{figure*}[h]
    \centering
   \includegraphics[width=0.88\linewidth]{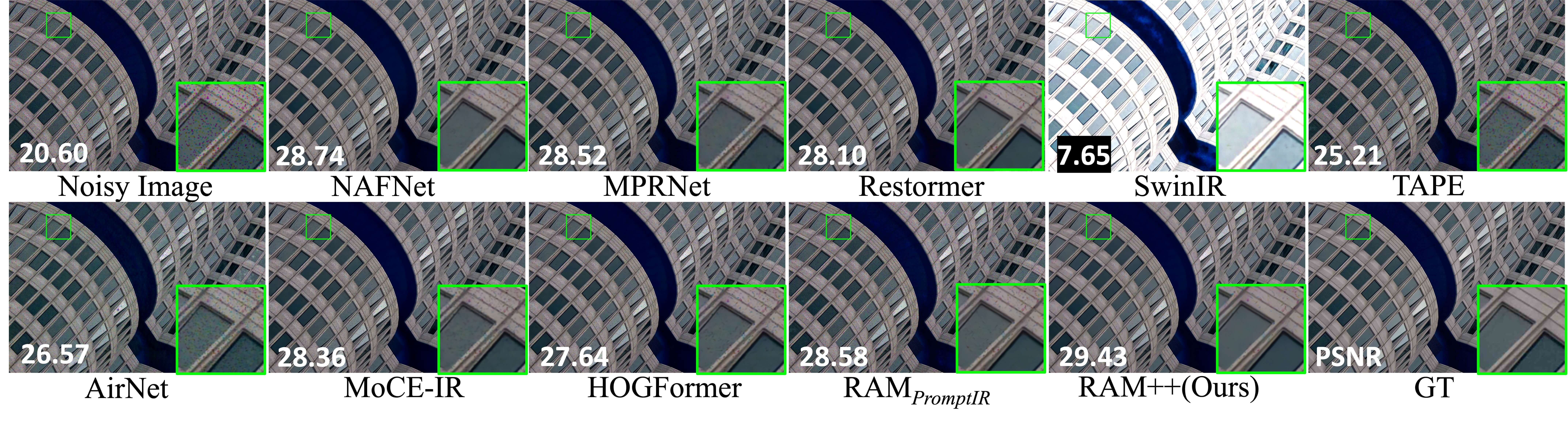}
   \caption{Denoise visual comparsion on OOD (Urban100-Pepper) dataset. Zoom in for details.}
   \label{fig:ood_result}
\end{figure*}

\textbf{Denoising.} As shown in Tab.~\ref{tab:denoise_test}, although only Gaussian noise is employed during the training of the seven tasks, our model achieves state-of-the-art performance for in-distribution noise and yields an average PSNR gain of $1.84$dB over the baseline (Restormer~\cite{Zamir2021Restormer}) on unseen noise types. These results highlight the model’s robustness and strong generalization across diverse degradations. Fig.~\ref{fig:ood_result} shows visual comparisons.

\begin{figure}[h]
  \centering
   \includegraphics[width=0.88\linewidth]{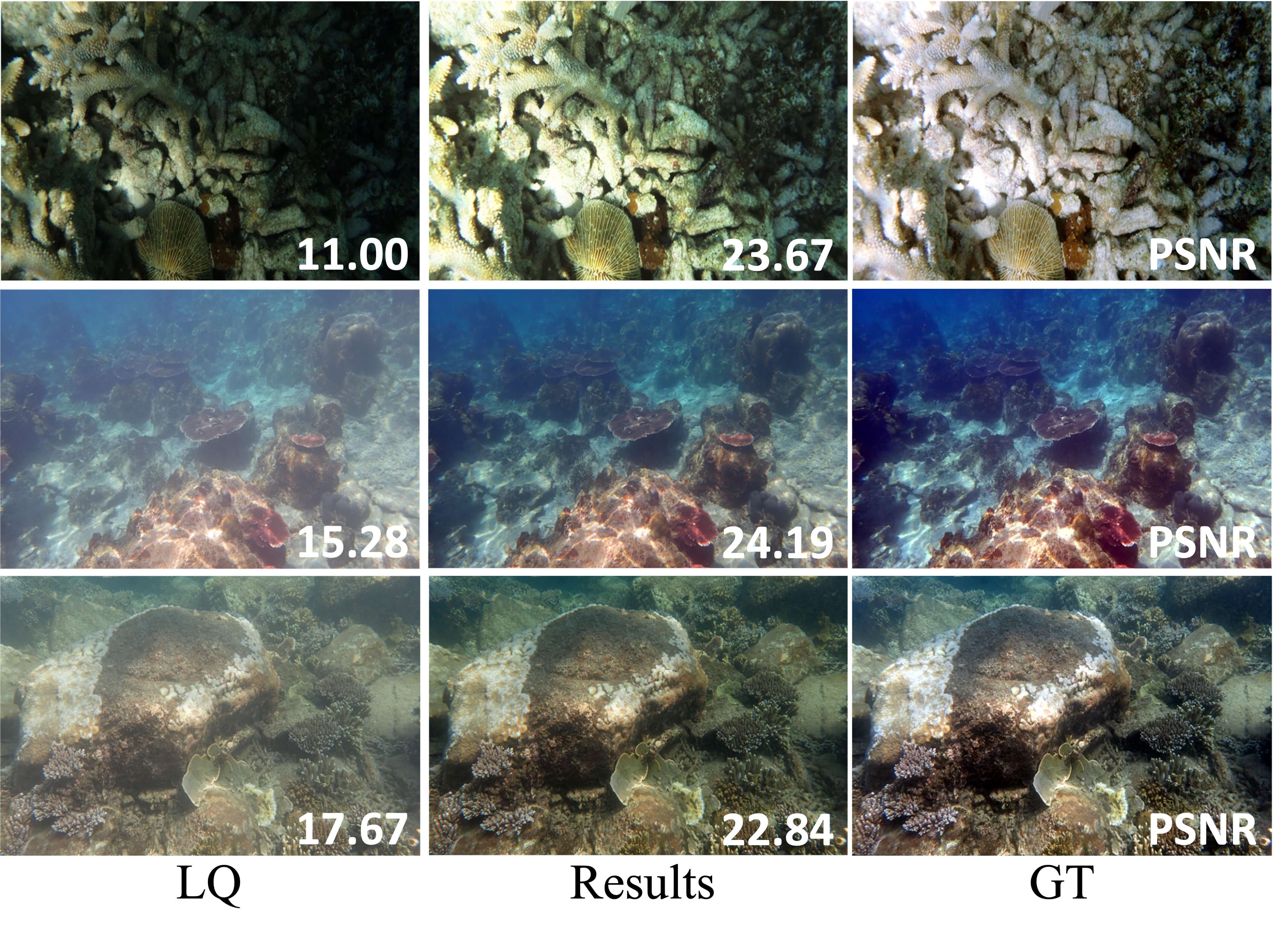}
   \caption{Underwater image enhancement visualization of our method on OOD (UIEB)  dataset. Zoom in for details.}
   \label{fig:uieb_result}
\end{figure}

\textbf{Underwater Enhancement} We follow the DCPT~\cite{hu2025universalimagerestorationpretraining} protocol and test the model trained under the 5-task setting. As shown in Tab.~\ref{tab:ood_uieb}, the results demonstrate not only highly consistent performance across in-distribution tasks but also strong generalization to OOD data. Fig.~\ref{fig:uieb_result} shows that our method leverages knowledge from related restoration tasks to effectively handle the color shift in underwater images.
% low-light enhancement and dehazing knowledge 
\begin{table*}[t]
\centering
\caption{UIEB generalization performance with five in-distribution dataset comparisons.}
\vspace{3pt} 
\resizebox{\textwidth}{!}{ % 缩放到版心宽度
\begin{tabular}{c|cccccc|c}
\toprule
 \multirow{2}{*}{Method}& \multicolumn{6}{c|}{In-Distribution} &  Out-of-Distribution \\
& SOTS~\cite{sots} &Ran100L~\cite{rain100} & BSD68~\cite{bsd68} & Gopro~\cite{gopro} & LOL~\cite{LOL} & Average &UIEB~\cite{UIEB} \\
\midrule
AirNet~\cite{airnet} & 21.04/0.884 & 32.98/0.951 & 30.91/0.882 & 24.35/0.781 & 18.18/0.735 & 25.49/0.846 & 15.46/0.745 \\
SwinIR~\cite{liang2021swinir} & 21.50/0.891 & 30.78/0.923 & 30.59/0.868 & 24.52/0.773 & 17.81/0.723 & 25.04/0.835 & 15.31/0.740 \\
NAFNet~\cite{nafnet} & 25.23/0.939 & 35.56/0.967 & 31.02/0.883 & 26.53/0.808 & 20.49/0.809 & 27.76/0.881 & 15.42/0.744 \\
Restormer~\cite{Zamir2021Restormer} & 24.09/0.927 & 34.81/0.962 & \textcolor{blue}{31.49}/0.884 & 27.22/0.829 & 20.41/0.806 & 27.60/0.881 & 15.46/0.745 \\
PromptIR~\cite{promptir} & 25.20/0.931 & 35.94/0.964 & 31.17/0.882 & 27.32/0.842 & 20.94/0.799 & 28.11/0.883 & 15.48/0.748 \\
DCPT$_{\mathrm{PromptIR}}$~\cite{hu2025universalimagerestorationpretraining} 
& \textcolor{teal}{30.72}/\textcolor{teal}{0.977} 
& 37.32/\textcolor{teal}{0.978} 
& 31.32/0.885 
& \textcolor{blue}{28.84}/\textcolor{blue}{0.877} 
& \textcolor{blue}{23.35}/0.840 
& \textcolor{teal}{30.31}/0.911 
& 15.78/\textcolor{teal}{0.772} \\
RAM$_{\mathrm{Restormer}}$~\cite{qin2024restore}&
30.22/0.976 & 
\textcolor{teal}{38.21}/\textcolor{blue}{0.983} & 
31.41/\textcolor{teal}{0.886} & 
28.36/0.866 & 
22.66/\textcolor{teal}{0.850} & 
30.17/\textcolor{teal}{0.912} & 
\textcolor{teal}{17.37}/\textcolor{blue}{0.780}\\
\methodname$_{\text{100\%}}$ 
& \textcolor{blue}{31.34}/\textcolor{blue}{0.979} 
& \textcolor{blue}{38.29}/\textcolor{blue}{0.983} 
& \textcolor{teal}{31.48}/\textcolor{blue}{0.887} 
& \textcolor{teal}{28.37}/\textcolor{teal}{0.867} 
& \textcolor{teal}{22.67}/\textcolor{blue}{0.852} 
& \textcolor{blue}{30.43}/\textcolor{blue}{0.914} 
& \textcolor{blue}{17.44}/\textcolor{blue}{0.780} \\
\bottomrule
\end{tabular}
}
\label{tab:ood_uieb}
\end{table*}

\begin{figure*}[h]
  \centering
   \includegraphics[width=0.88\linewidth]{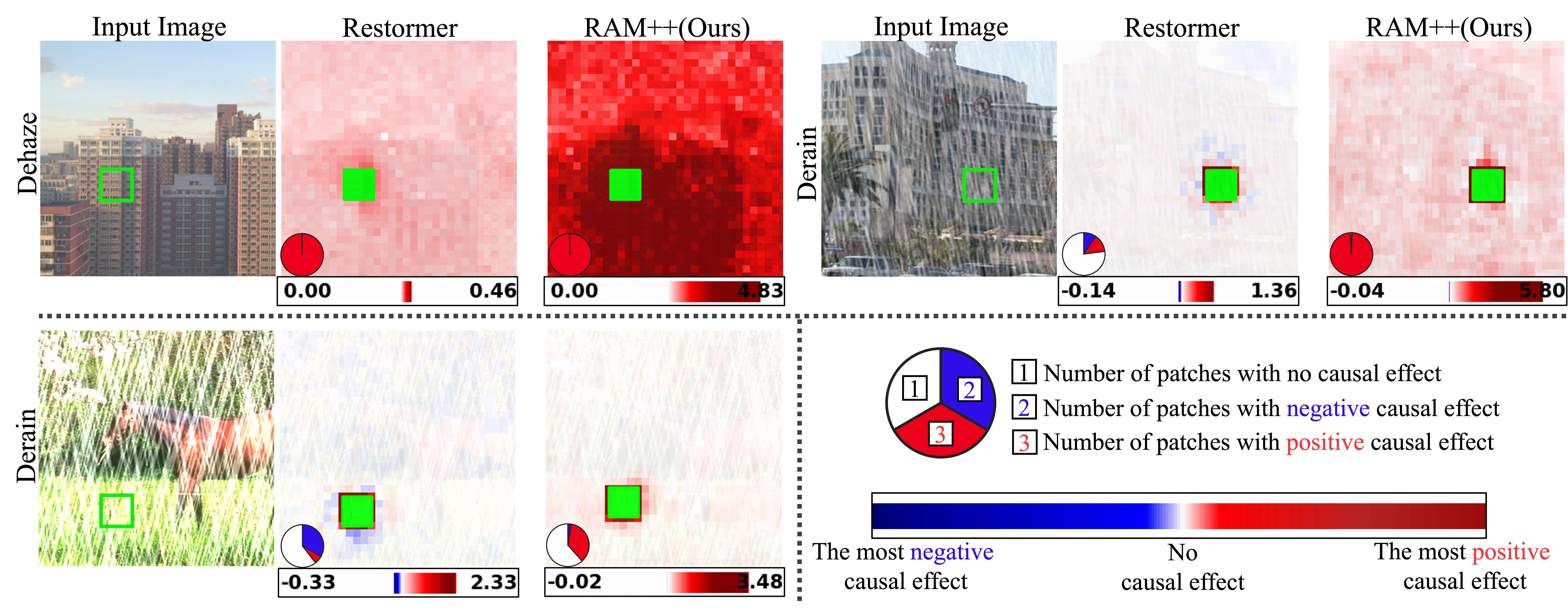}
       \caption{CEMs in image dehazing and deraining tasks. Patches with positive or negative causal effects and the regions of interest (ROIs) are indicated in red, blue, and green, respectively. The color intensity denotes the magnitude of the effect. The colorbar denotes the causal effect range, and the pie chart shows the relative fractions of positive (red) and negative (blue) patches. The first-row image highlights architecture as the main subject, whereas the second-row image presents multiple texture-rich subjects. The results indicate that our method is capable of maintaining a broad receptive field, precisely attending to positive contribution regions, and effectively suppressing misleading signals from irrelevant areas.}
   \label{fig:cem}
\end{figure*}

\subsection{Analysis of Performance Enhancement}
To improve the interpretability of our method, we analyze the underlying causes of its improvements in in-distribution performance and out-of-distribution generalization.
\paragraph{Effective Semantic Understanding}
Prior study~\cite{jin2025classic} suggests that semantic understanding is the cornerstone of image restoration. A model must first capture the semantic content of an image before it can reliably separate degradations from authentic textures. Many existing all-in-one image restoration approaches neglect this intrinsic semantic learning, which constrains their ability to handle the intricate interplay between degradations and original structures. Fig.~\ref{fig:rain_result} shows that, while the previous method (\eg, MoCE-IR~\cite{moceir}) successfully removes most rain streaks outside the subject, residual streaks remain on facial regions. In contrast, our method explicitly emphasizes intrinsic image representation, enabling more sensitive and comprehensive detection of degradations and yielding natural, high-quality restorations. Remarkably, in scenarios where rain streaks are strongly coupled with the intrinsic textures of the image, our approach is the only one that does not fail.
\paragraph{Stable Global Information Acquisition}
In the field of low-level vision, state-of-the-art models, with Restormer~\cite{Zamir2021Restormer} as a representative, typically adopt a global receptive field in the hope of capturing features across a broader pixel range.
However, prior work~\cite{cem} has revealed that when unified restoration models are trained in multi-task settings, they often degenerate into over-reliance on local neighborhoods, thereby losing the ability to integrate long-range knowledge, \ie, expanding the receptive field does not necessarily translate into effective utilization of more pixel information~\cite{lam}. 
We employ the Causal Effect Map (CEM)\cite{cem} to validate that our method strives to maintains stable global information acquisition even under multi-task training. 
CEM elucidates the causal relationship between input patches and the Region Of Interest (ROI) in the output image, highlighting the patches exerting causal influence—depicted in red for positive effects and blue for negative effects.
Fig.~\ref{fig:cem}(first row) demonstrates that \methodname{} consistently leverages positive global contextual information when the ROI lies within the salient image subject, irrespective of whether for long-range modeling tasks (\eg, dehazing) or local detail recovery (\eg, deraining).
\begin{table*}[h]
  \centering
  \setlength{\tabcolsep}{8pt}
  \small
  \caption{Ablative results of various components and strategies.}
  \vspace{3pt} 
  \label{tab:combined_ablation}

  % 第一行 两个表格
  \noindent
  \begin{minipage}[t]{0.45\textwidth}
    \centering
    (a) Ablative results of different components.
    \vspace{0.5em}
    \begin{tabular}{c|cc}
      \toprule
      \methodname & PSNR$\uparrow$ & SSIM$\uparrow$ \\ \midrule
      w/o AdaSAM\&RFR & 26.01 & 0.8007 \\
      w/o AdaSAM & 28.40 & 0.8811 \\
      w/o RFR & 28.55 & \textcolor{blue}{0.8897} \\
      w/ AdaSAM\&RFR (Ours) & \textcolor{blue}{28.88} & 0.8895 \\
      \bottomrule
    \end{tabular}
  \end{minipage}
  \hspace{0.04\textwidth}
  \begin{minipage}[t]{0.45\textwidth}
    \centering
    (d) Ablative results of different fine-tuning strategies.

    \vspace{0.5em}

    \begin{tabular}{c|cc}
      \toprule
      Method & PSNR$\uparrow$ & SSIM$\uparrow$ \\ \midrule
      random & 28.40 & 0.8868 \\
      IG~\cite{ig} & 28.48 & 0.8882 \\
      MAC (Ours) & \textcolor{blue}{28.88} & \textcolor{blue}{0.8895} \\
      \bottomrule
    \end{tabular}
  \end{minipage}

  \vspace{0.8em}

  % 第二行 两个表格
  \noindent
  \begin{minipage}[t]{0.45\textwidth}
    \centering
   (b) Ablative results of different pre-training strategies.

    \vspace{0.5em}

    \begin{tabular}{c|cc}
      \toprule
      Method & PSNR$\uparrow$ & SSIM$\uparrow$ \\ \midrule
      pre-trained w/ gt & 27.58 & 0.8740 \\
      pre-trained w/ paired data & \textcolor{blue}{28.88} & \textcolor{blue}{0.8895} \\
      \bottomrule
    \end{tabular}
  \end{minipage}
  \hspace{0.04\textwidth}
  \begin{minipage}[t]{0.45\textwidth}
    \centering
   (e) Ablative results of different feature-level design.

    \vspace{0.5em}

    \begin{tabular}{c|cc}
      \toprule
      Method & PSNR$\uparrow$ & SSIM$\uparrow$ \\ \midrule
      Single-level & 28.02 & 0.8843 \\
      Multi-level (Ours) & \textcolor{blue}{28.88} & \textcolor{blue}{0.8895} \\
      \bottomrule
    \end{tabular}
  \end{minipage}

  \vspace{0.8em}

  % 第三行 两个表格
  \noindent
  \begin{minipage}[t]{0.45\textwidth}
    \centering
  (c) Ablative results of different mask strategies.

    \vspace{0.5em}
    \begin{tabular}{c|cc}
      \toprule
      Method & PSNR$\uparrow$ & SSIM$\uparrow$ \\ \midrule
      Random$_{8*8}$ & 28.89 & 0.8967 \\
      Random$_{1*1}$ & 29.01 & 0.8968 \\
      AdaSAM & \textcolor{blue}{29.46} & \textcolor{blue}{0.8993} \\ \bottomrule
    \end{tabular}
  \end{minipage}
  \hspace{0.04\textwidth}
  \begin{minipage}[t]{0.45\textwidth}
    \centering
   (f) Ablative results of different feature fusion strategies.

    \vspace{0.5em}
    \begin{tabular}{c|cc}
      \toprule
      Method & PSNR$\uparrow$ & SSIM$\uparrow$ \\ \midrule
      Cross-attention & 28.82 & 0.8947 \\
      SFT~\cite{sft} & 28.82 & 0.8955 \\
      Ours & \textcolor{blue}{29.46} & \textcolor{blue}{0.8993} \\
      \bottomrule
    \end{tabular}
  \end{minipage}

  \vspace{0.8em}

  % 你之前的那个大表格，如果想，也可以放这里
    \setlength{\tabcolsep}{3pt}
   (g) Ablative results of different fine-tuning ratios.
   \vspace{0.5em} 
   
    \resizebox{\textwidth}{!}{ % 缩放到版心宽度
    \begin{tabular}{c|cccc|ccccc}
      \toprule
      \multirow{2}{*}{Method} & \multicolumn{4}{c|}{In-Distribution} & \multicolumn{5}{c}{Out-Of-Distribution} \\
      & $\sigma=15$ & $\sigma=25$ & $\sigma=50$ & Average$\uparrow$ & Poisson & Pepper & Speckle & Average$\uparrow$ & SRGA$\downarrow$ \\ \midrule
      \methodname$_{10\%}$ & 33.73/0.940 & 31.50/0.913 & 28.33/0.855 & 31.19/0.903 & 11.32/0.193 & 25.09/0.788 & 16.48/0.461 & \textcolor{blue}{17.63}/\textcolor{blue}{0.480} & \textcolor{blue}{2.50} \\
      \methodname$_{30\%}$ & 33.87/0.942 & 31.63/0.915 & 28.49/0.859 & 31.33/0.905 & 11.04/0.178 & 24.84/0.776 & 15.50/0.423 & 17.13/0.459 & 2.58 \\
      \methodname$_{50\%}$ & 33.98/0.942 & 31.72/0.916 & 28.57/0.860 & 31.42/0.906 & 10.79/0.175 & 24.54/0.766 & 15.36/0.418 & 16.89/0.453 & 2.64 \\
      \methodname$_{100\%}$ & 34.15/0.944 & 31.87/0.918 & 28.70/0.863 & \textcolor{blue}{31.57}/\textcolor{blue}{0.908} & 10.34/0.157 & 23.94/0.749 & 14.92/0.385 & 16.40/0.430 & 2.84 \\
      \bottomrule
    \end{tabular}
  }
\end{table*}
\paragraph{Accurate Positive/Negative Information Discrimination}
However, expanding the receptive field is not always beneficial, as an excessively wide scope often introduces irrelevant information and may even impair restoration, making it crucial to distinguish between positive and negative regions. 
As shown in Fig.~\ref{fig:cem}(first row), in dehazing task, the most positively influential patch in Restormer~\cite{Zamir2021Restormer} yields only a $0.46$dB improvement in PSNR, whereas in deraining task, its most detrimental patch causes a $0.14$dB drop. 
In contrast, \methodname{} achieves up to a $4.83$dB gain while suffering only a $0.04$dB decline. 
Moreover, as shown in Fig.~\ref{fig:cem}(second row), the scene depicts a complex composition with multiple textured objects (\eg, grass, trees, and horse). 
As the ROI falls on only one of them, both models exploit only about $40$\% of the image pixels. 
However, among the pixels utilized by Restormer\cite{Zamir2021Restormer}, as many as $84$\% exert negative contributions (mainly from the upper trees), whereas \methodname{} reduces this proportion to merely $7$\%, with most positive contributions concentrated around the ROI region, which is consistent with the local nature of the deraining task.
These findings highlight that our model not only exhibits strong robustness against negative information but also makes more effective use of positive contributions.
% \paragraph{Focusing on Positive Contribution Regions}
% The key challenge of image restoration is to capture regions that genuinely benefit the target area within a broad receptive field, while avoiding interference from irrelevant information~\cite{lam}. To this end, we employ the Causal Effect Map (CEM)\cite{cem} to quantify the positive and negative contributions of different input regions to the ROI. As shown in Fig.~\ref{fig:cem}a, b, when the ROI lies in the main part of the image, whether for long-range modeling tasks (\eg, dehazing) or local detail recovery (\eg, deraining), our restoration model leverages the global receptive field to extract semantically relevant context.

% However, expanding the receptive field is not always beneficial, as an excessively wide scope often introduces irrelevant information and may even impair restoration. Fig.~\ref{fig:cem}c shows that when the ROI lies in background regions such as land and lacks semantic correlation with other elements (\eg, person, benches), our model does not blindly enlarge the utilized pixel range. Instead, it focuses on semantically relevant regions that provide genuine positive contributions (e.g., adjacent land), thereby effectively avoiding interference.

% In summary, by focusing on positively contribution regions within the global receptive field, the model efficiently leverages useful context while suppressing distractions, yielding more robust restoration.

\paragraph{Prioritized Background Structure Reconstruction}
During training, restoration models often adopt one of two strategies to minimize loss~\cite{generalizationlowlevel}: (i) identifying and removing degradations, or (ii) recognizing and reconstructing background structures. Since multi-degradation datasets provide abundant background information, learning to remove a fixed number or pattern of degradations (strategy i) is usually simpler, which causes networks to overfit to degradation removal. By contrast, strategy (ii) naturally promotes better generalization, as it requires learning more comprehensive representations. Our pre-training and fine-tuning scheme compels the model to prioritize background reconstruction, thereby substantially enhancing its ability to generalize beyond the training distribution.

\subsection{Ablation Study}
\label{sec:ablation}
In this section, we conduct ablation studies on the pre-training strategy, masking strategy, fine-tuning strategy, fine-tuning ratio, feature selection and fusion strategies to demonstrate the effectiveness of our method.

\paragraph{Different Components}
To verify the contributions of each core component in our method, we design the following three experimental configurations:
In the first configuration, we remove AdaSAM; In the second configuration, AdaSAM is introduced, but the regularization module RFR is removed; The third configuration corresponds to our full proposed method, which integrates both AdaSAM and RFR.

As shown in Tab.~\ref{tab:combined_ablation}(a), using only the AdaSAM pretraining strategy can guide the network to learn content-aware representations. However, without the structured regularization from RFR, the model struggles to fully leverage the pretrained features, resulting in limited performance. By further introducing the RFR module based on DINOv2 features, this issue is effectively alleviated, and the semantic consistency is enhanced, enabling the model to achieve the best performance.

\paragraph{Pre-training with Paired Data}
Tab.~\ref{tab:combined_ablation}(b) compares our paired-data MIM pre-training strategy with the variant that uses only ground truth images. The results show that pre-training with paired data is essential for the effectiveness of \methodname.
Using only high-quality images for pre-training does not effectively guide the model to learn image restoration. The presence of paired data is necessary to enable meaningful learning.

\paragraph{Masking Strategy}
We first determine the mask patch size, an essential hyperparameter that governs the continuity and area of masked regions in an image.
In high-level vision tasks, MAE~\cite{mae} masks $75\%$ of an image using a $16\times16$ patch size. However, such a large patch size can severely corrupt local image details, making it unsuitable for image restoration.
To investigate the optimal patch size, we pre-train SwinIR~\cite{liang2021swinir} with patch sizes of $1\times1$, $4\times4$, and $8\times8$, as illustrated in Fig.~\ref{fig:patch-size}. Since SwinIR’s attention layers treat an $8\times8$ patch as a token, pre-training with $4\times4$ patches introduces severe artifacts. Conversely, using $8\times8$ patches during pre-training produces outputs that lack fine details, such as the texture on a polar bear’s paws. In contrast, the model pre-trained with a $1\times1$ patch size achieves higher reconstruction quality and effectively removes most rain streaks. Therefore, AdaSAM ultimately adopts a pixel-level masking scheme.

To fully leverage the knowledge learned during the pre-training stage, we fine-tune three pre-trained models under a full ($100\%$) fine-tuning setup.
As shown in Tab.~\ref{tab:combined_ablation}(c), AdaSAM surpasses random masking with an $8\times8$ patch size by $0.57$dB in PSNR and improves upon a fixed $1\times1$ patch size by $0.45$dB, clearly demonstrating the effectiveness of the Adaptive Semantic-Aware Mask.
\begin{figure*}[]
  \centering
   \includegraphics[width=0.85\linewidth]{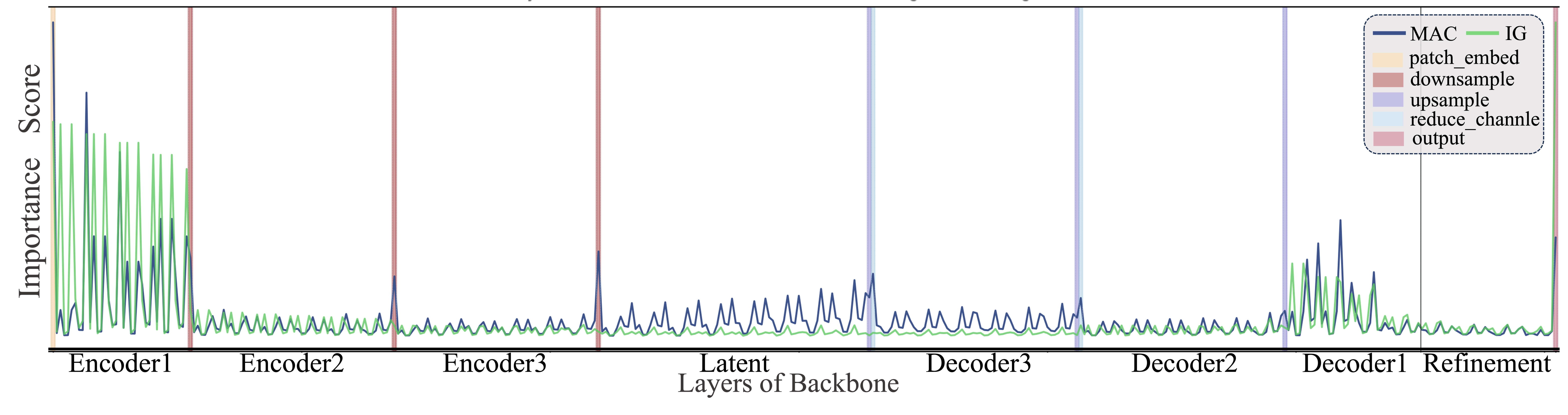}
   \caption{Comparative importance analysis of MAC and IG. Our analysis emphasizes the intrinsic image information in the latent layer.}
   \label{fig:ablation_mac}
\end{figure*}

\begin{figure*}[h]
  \centering
   \includegraphics[width=0.88\linewidth]{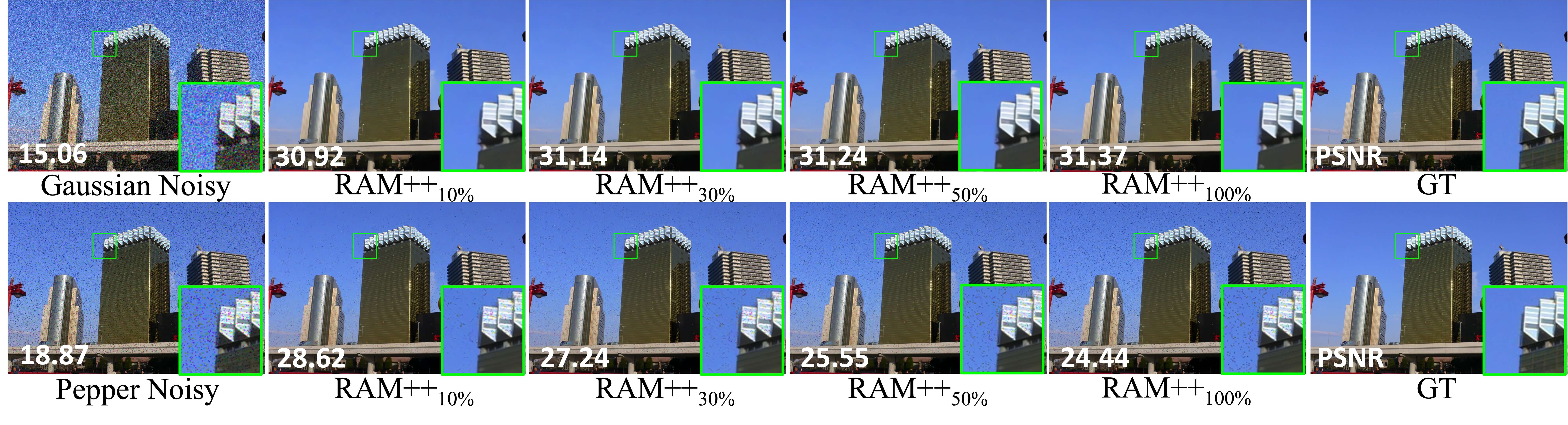}
   \caption{Fine-tuning ratio visual comparison on Urban dataset. Gaussian noise (ID) and Pepper noise (OOD) are used for evaluation. Zoom in for details.}
   \label{fig:ablation_ratio}
\end{figure*}

\paragraph{Fine-tuning Strategy}
To evaluate the effectiveness of our fine-tuning strategy, we fine-tune $30\%$ of the network layers selected using MAC analysis, IG~\cite{ig}, and uniform sampling.
As shown in Tab.~\ref{tab:combined_ablation}(d), our method outperforms IG by $0.40$dB in PSNR, demonstrating the superiority of our layer selection strategy. Fig.~\ref{fig:ablation_mac} illustrates that the patch embedding, the shallow layers of both the encoder and decoder, as well as the downsampling layers, primarily compensate for input incompleteness. Notably, MAC places greater emphasis on the contribution of the deepest latent representations than IG, aligning well with our objective of capturing the essential information in images. 
% Moreover, within each Transformer block, the terminal Gated-Dconv Feed-Forward Network(GDFN)  module and the second LayerNorm emerge as the most influential components.

\paragraph{Multi-level Representations}
We compare the performance of multi-level weighted feature fusion against single-level feature(\eg, single deep-layer feature) extraction to evaluate its effectiveness in enhancing image restoration quality. Tab.~\ref{tab:combined_ablation}(e) shows the multi-level design yields a PSNR gain of $0.86$dB, demonstrating its effectiveness.

\paragraph{Feature Fusion Strategy}
Although we attribute the performance improvement to the pre-training strategy and the strong inherent robust priors of DINOv2 rather than to a specific feature fusion strategy, we still compare our feature fusion strategy with two widely adopted baselines—Cross-Attention and Spatial Feature Transform (SFT)~\cite{sft}—under a full ($100\%$) fine-tuning setup.
Tab.~\ref{tab:combined_ablation}(f) shows our method achieves the best performance, surpassing SFT by $0.64$dB in PSNR, demonstrating the advantage of our fusion design.

\paragraph{Fine-tuning Ratio}
Since it is difficult to define a strictly out-of-distribution degradation for the model fine-tuned on seven types of degradation, we perform fine-tuning using Gaussian noise on different fine-tuning ratios. The results are shown in Tab.~\ref{tab:combined_ablation}(g).
With our strategy, fine-tuning only a small portion of the network (\eg, $10\%$) yields performance comparable to full fine-tuning.
However, to achieve the best task-specific performance, tuning most or all parameters is still required.

\paragraph{Performance \textit{vs} Generalization Capability}
SRGA is a non-parametric metric designed to assess the generalization capability of models without the need for additional training~\cite{Generalization_Ability}. A lower SRGA value indicates better generalization ability.
In Tab.~\ref{tab:combined_ablation}(g), we observe a trade-off between in-distribution performance and out-of-distribution generalization.
The more layers are fine-tuned, the more the model tends to overfit to the training distribution, reducing its generalization ability.
Fig.~\ref{fig:ablation_ratio} shows that our fine-tuning approach enables the model to retain strong generalization while maintaining comparable performance.

\section{Conclusion, Limitation, and Future Work}
% This paper presents \methodname{}, a pipeline for extracting intrinsic image information from degraded images using Mask Image Modeling (MIM) pre-training. 
% We design a MIM pre-training strategy tailored for image restoration and a fine-tuning algorithm to handle the transition from masked to complete images.
% By analyzing layer importance with MAC, we achieve high performance with minimal parameter tuning.
% Extensive experiments demonstrate that our \methodname{} can bring boosts to various architectures and achieve state-of-the-art performance, moving towards a unified solution for all-in-one image restoration.
This paper presents \methodname{}, a pipeline designed to extract robust intrinsic
information from degraded images with genuine potential for \textit{all-in-one} image restoration, transcending the limitations of \textit{multi-in-one} approaches. Through Adaptive Semantics-Aware Mask image modeling, MAC-guided layer-wise fine-tuning, and DINOv2-driven Robust Feature Regularization, \methodname{} delivers robust, balanced, and state-of-the-art results across diverse degradations. Notably, its advantages amplify as task count grows, underscoring its promise as a foundation for future all-in-one image restoration systems.

Despite its effectiveness, \methodname{} has limitations. Fine-tuning on a mixed dataset of diverse degradations inevitably faces inherent conflicts between tasks, and the characteristics of masked image modeling can make inferring details for kernel deblurring more challenging. Future work includes multi-task learning and optimized data-mixing strategies, as well as extending the framework to video restoration with temporal consistency to strengthen its general-purpose restoration capability.

%\section*{Acknowledgments}
%This should be a simple paragraph before the References to thank those individuals and institutions who have supported your work on this article.

%{\appendix[Proof of the Zonklar Equations]
%Use $\backslash${\tt{appendix}} if you have a single appendix:
%Do not use $\backslash${\tt{section}} anymore after $\backslash${\tt{appendix}}, only $\backslash${\tt{section*}}.
%If you have multiple appendixes use $\backslash${\tt{appendices}} then use $\backslash${\tt{section}} to start each appendix.
%You must declare a $\backslash${\tt{section}} before using any $\backslash${\tt{subsection}} or using $\backslash${\tt{label}} ($\backslash${\tt{appendices}} by itself
% starts a section numbered zero.)}



\bibliographystyle{IEEEtran}

\begin{IEEEbiography}
[{\includegraphics[width=1in,height=1.25in,clip,
keepaspectratio]{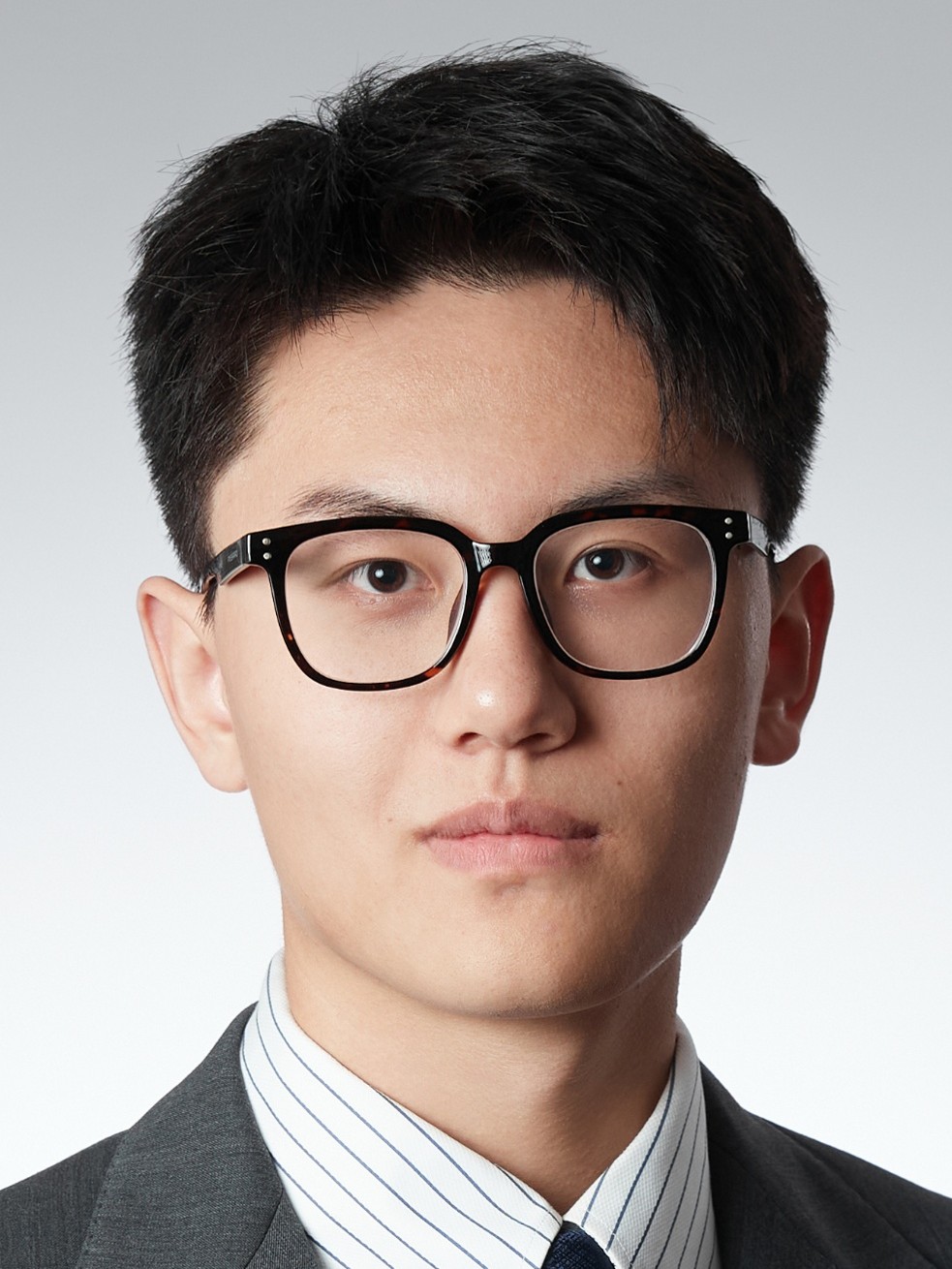}}]
{Zilong Zhang}
is currently a Ph.D. student at the College of Computer Science, Nankai University, supervised by Prof. Chongyi Li. He received a B.S. degree in Engineering from Hainan University in 2025. His research interests focus on low-level computer vision.
\vspace{-4mm}
\end{IEEEbiography}

\begin{IEEEbiography}
[{\includegraphics[width=1in,height=1.25in,clip,
keepaspectratio]{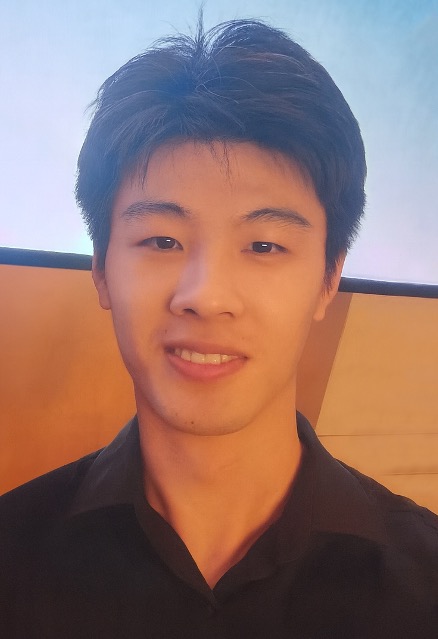}}]
{Chujie Qin}
received his B.S. degree from the
College of Computer Science, Nankai University,
China, in 2023. He is currently a Ph.D. student
at the College of Computer Science, Nankai
University. His research interests include computational photography, unified image restoration.
\vspace{-4mm}
\end{IEEEbiography}

\begin{IEEEbiography}
[{\includegraphics[width=1in,height=1.25in,clip,
keepaspectratio]{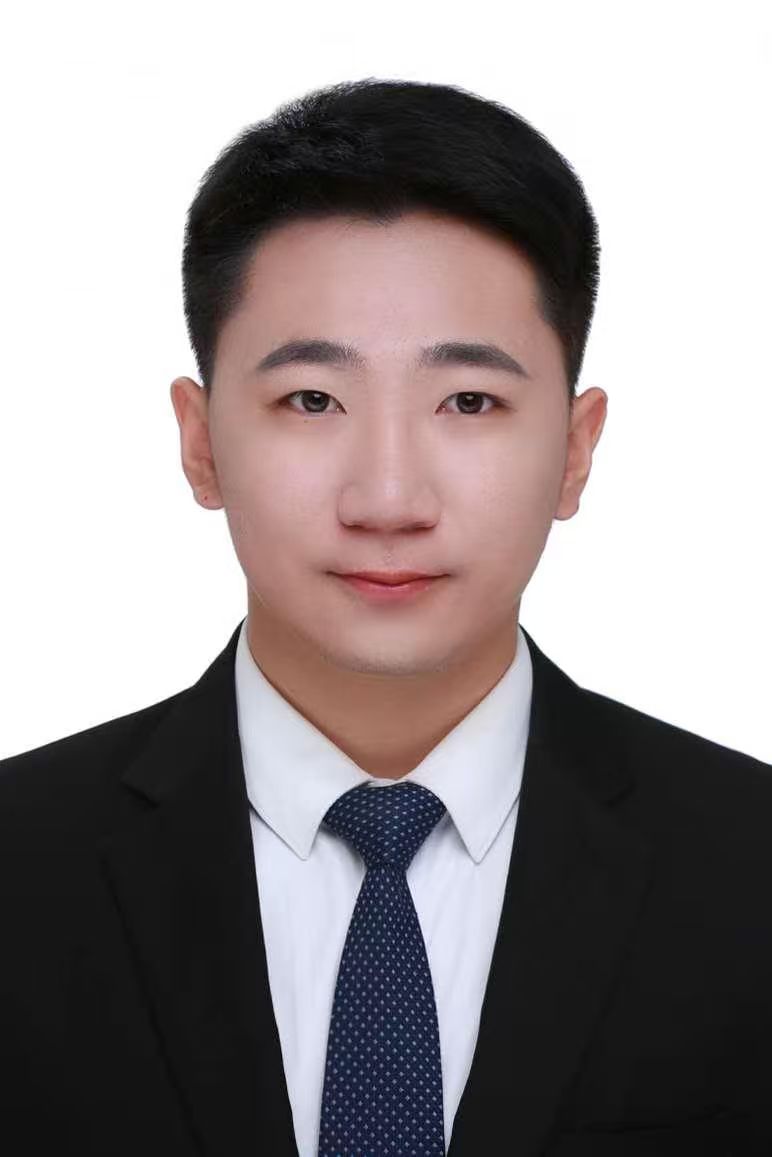}}]
{Chunle Guo}
 (Member, IEEE) received a Ph.D. degree from Tianjin University, China, under the supervision of Prof. Jichang Guo. He was a Visiting Ph.D. Student with the School of Electronic Engineering and Computer Science, Queen Mary University of London (OMUL), U.K. He was a Research Associate with the Department of Computer Science, City University of Hong Kong (CityU). He was a Postdoctoral Researcher with Prof. Ming-Ming Cheng at Nankai University. He is currently an Associate Professor at Nankai University. His research interests include image processing, computer vision, and deep learning.
\vspace{-4mm}
\end{IEEEbiography}

\begin{IEEEbiography}
[{\includegraphics[width=1in,height=1.25in,clip,
keepaspectratio]{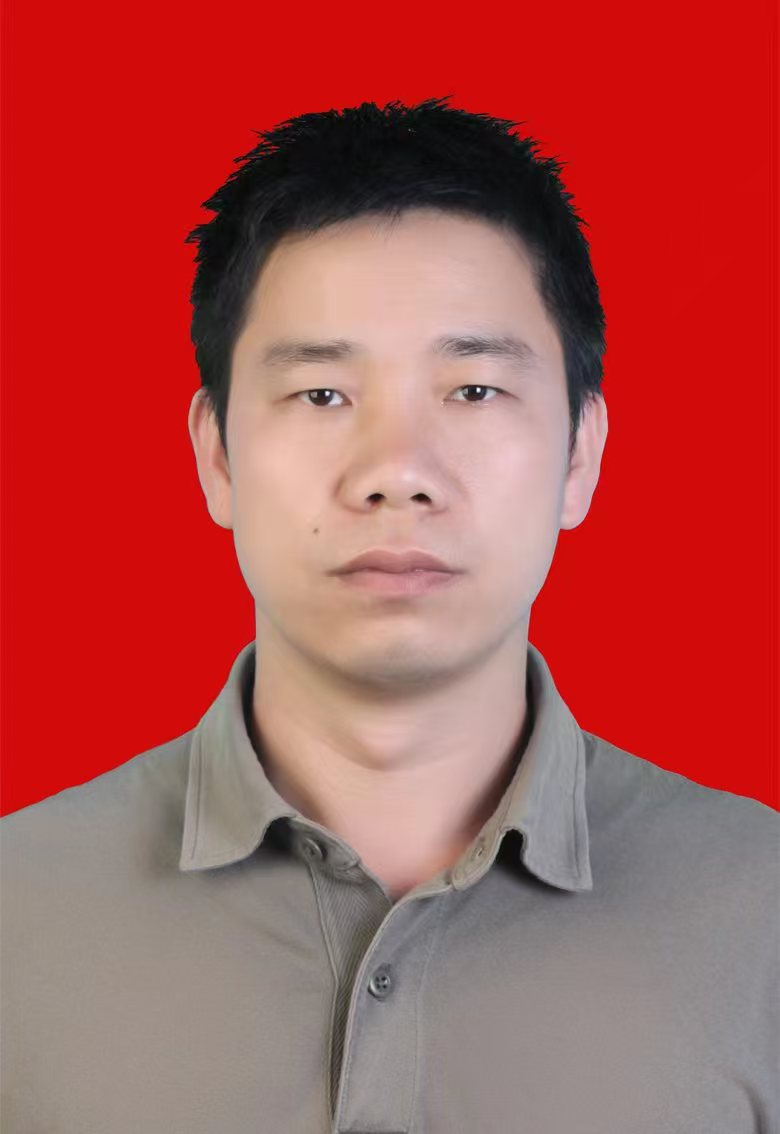}}]
{Yong Zhang}
works as a professional engineer at Chongqing Chang'an Industrial (Group) Co., Ltd. Shenzhen Branch, mainly engaged in system integration and software development. He is also an in-service doctoral student in the Multimedia Laboratory of the School of Computer Science at Nankai University. His main research directions include object detection and tracking, multimodal perception, and data fusion.
\vspace{-4mm}
\end{IEEEbiography}

\begin{IEEEbiography}
[{\includegraphics[width=1in,height=1.25in,clip,
keepaspectratio]{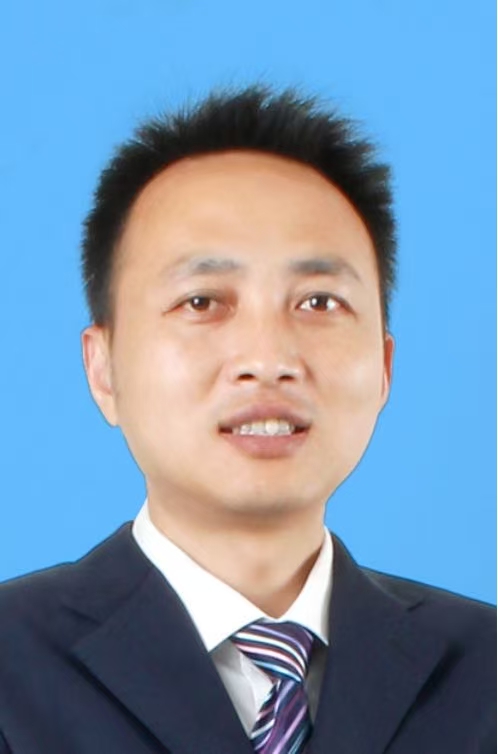}}]
{Chao Xue}
received the B.S. degree in automation from Tianjin University, Tianjin, China, in 2002. He is currently the Senior Deputy General Manager and  the Chief Technology Officer at Tiandy Technologies. He is a Professor-Level Senior Engineer. He is also an Executive Director of Tianjin Optical Society, the Director of Tianin industrial Design Association, and the Director of Tianiin Association for Artificial Intelligence. His research interests include video and image processing, intelligent analysis, and security business application.
\vspace{-4mm}
\end{IEEEbiography}

\begin{IEEEbiography}
[{\includegraphics[width=1in,height=1.25in,clip,
keepaspectratio]{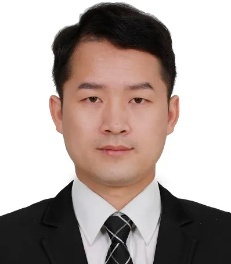}}]
{Ming-Ming Cheng}
 received his PhD degree from Tsinghua University in 2012. Then, he did 2 years research fellow, with Prof. Philip Torr in Oxford. He is now a professor at Nankai University, leading the Media Computing Lab. His research interests include computer graphics, computer vision, and image processing. He received research awards, including the National Science Fund for Distinguished Young Scholars and the ACM China Rising Star Award. He is on the editorial boards of IEEE TPAMI and IEEE TIP.
\vspace{-4mm}
\end{IEEEbiography}

\begin{IEEEbiography}
[{\includegraphics[width=1in,height=1.25in,clip,keepaspectratio]{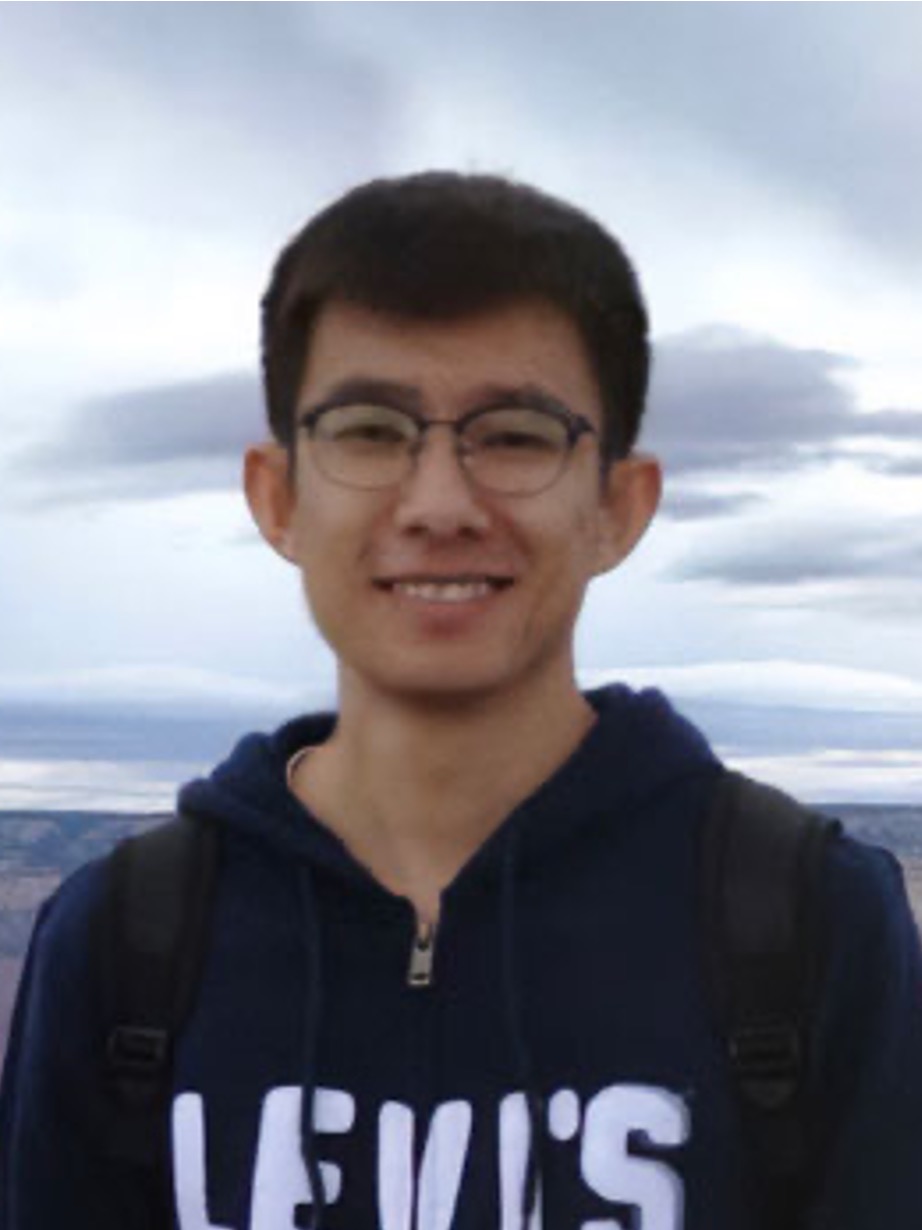}}]
{Chongyi Li} is a Professor with the School of Computer Science, Nankai University, China.  He was a  Research Assistant Professor with the S-Lab,  School of Computer Science and Engineering, Nanyang Technological University, Singapore from 2021 to 2023.  He was a Research Fellow with the City University of Hong Kong and Nanyang Technological University from 2018 to 2021. He received his Ph.D. degree from Tianjin University in 2018. He was also a joint training Ph.D. student at the Australian National University, Australia, from 2016 to 2017. His research interests include computer vision, machine learning, and computational imaging, particularly in image enhancement and restoration, image generation and editing, and underwater imaging. 
\vspace{-4mm}
\end{IEEEbiography}

%\newpage

 

% \end{IEEEbiography}



% \vfill

\end{document}